\documentclass[12pt,a4paper]{tesis}

\usepackage{graphicx}
\usepackage[utf8]{inputenc}
\usepackage[left=3cm,right=3cm,bottom=3.5cm,top=3.5cm]{geometry}

\usepackage[section]{placeins}
\usepackage{filecontents}
\usepackage{caption}
\usepackage{subcaption}
\usepackage{siunitx}
\usepackage{bm}
\usepackage{multirow}
\usepackage{amsmath}
\usepackage[breaklinks=true,bookmarks=false]{hyperref}
\usepackage{listings}
\usepackage{xcolor}
\usepackage{comment}
\usepackage{csquotes}

\newcommand{\etal}{\textit{et al.}}

\begin{document}

%%%% CARATULA

\def\autor{Juli\'an Del Gobbo}
\def\tituloTesis{Unconstrained Text Detection in Manga}
\def\runtitle{Unconstrained Text Detection in Manga}
\def\runtitulo{Detecci\'on de Texto sin Restricciones en Manga}
\def\director{Rosana Matuk Herrera}
\def\lugar{Buenos Aires, 2020}
\newcommand{\HRule}{\rule{\linewidth}{0.2mm}}
\thispagestyle{empty}

\begin{center}\leavevmode

\vspace{-2cm}

\begin{tabular}{l}
\includegraphics[width=2.6cm]{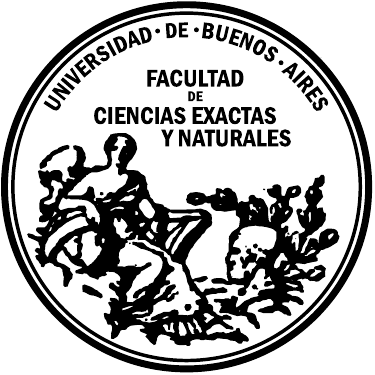}
\end{tabular}

{\large \sc Universidad de Buenos Aires

Facultad de Ciencias Exactas y Naturales

Departamento de Computaci\'on}

\vspace{6.0cm}

%\vspace{3.0cm}
%{
%\Large \color{red}
%\begin{tabular}{|p{2cm}cp{2cm}|}
%\hline
%& Pre-Final Version: \today &\\
%\hline
%\end{tabular}
%}
%\vspace{2.5cm}

\begin{huge}
\textbf{\tituloTesis}\par
\end{huge}

\vspace{2cm}

{\large Tesis de Licenciatura en Ciencias de la Computaci\'on}

\vspace{2cm}

{\Large \autor}

\end{center}

\vfill

{\large

{Directora: \director}

\vspace{.2cm}

\lugar
}

\newpage\thispagestyle{empty}

\frontmatter
\pagestyle{empty}
\begin{center}
\large \bf \runtitle
\end{center}

\noindent The detection and recognition of unconstrained text is an open problem in research. 
Text in comic books has unusual styles that raise many challenges for text detection.
This work aims to identify text characters at a pixel level in a comic genre with highly sophisticated text styles: Japanese manga. To overcome the lack of a manga dataset with individual character level annotations, we create our own. Most of the literature in text detection use bounding box metrics, which are unsuitable for pixel-level evaluation. Thus, we implemented special metrics to evaluate performance. Using these resources, we designed and evaluated a deep network model, outperforming current methods for text detection in manga in most metrics.
\bigskip

\noindent\textbf{Keywords:} text-segmentation, datasets and evaluation, neural-networks, Japanese-text-detection, manga

\vspace{2cm}

\begin{center}
\large \bf \runtitulo
\end{center}

\noindent La detección y reconocimiento de texto sin restricciones es un problema abierto en la investigación. El texto en comics presenta estilos inusuales que plantean muchos desafíos para su detección. Este trabajo apunta a identificar caracteres de texto a nivel de píxel en un género de comics con estilos de texto muy sofisticados: el manga Japonés. Para superar la falta de dataset de manga con anotaciones por caracter, creamos nuestro propio. La mayoría de la literatura en detección de texto utiliza métricas basadas en coordenadas de rectángulos contenedores, los cuales son inadecuados para evaluar a nivel de píxel. Entonces, implementamos métricas especiales para evaluar el desempeño. Usando estos recursos, diseñamos y evaluamos un modelo de redes neuronales profundas, superando métodos actuales de detección de texto en manga en la mayoría de las métricas. 
\bigskip

\noindent\textbf{Palabras claves:} segmentación-de-texto, datasets y evaluación, redes-neuronales, detección-de-texto-japonés, manga

\clearpage

\begin{center}
\large \bf Agradecimientos
\end{center}

A Rosana, por acompa\~narme todo este tiempo y seguir luchando hasta que quedara lo m\'as perfecto posible. Agradezco su fuerte compromiso y las numerosas veces que nos juntamos. Gracias a todo ese esfuerzo finalmente logramos publicar.
A Daniel y Enrique, por tomarse el tiempo de leer la tesis y ser mi jurado. 
A exactas, por todo lo que me ense\~n\'o y aport\'o todos estos a\~nos.
A mis compa\~neros con quienes compart\'i maravillosas cursadas. 
A mis amigos, con quienes compart\'i muchas experiencias.
A mis compa\~neros de trabajo, que son como otra familia que me acompa\~n\'o y apoy\'o desde antes de empezar la carrera, especialmente Marcelo que siempre fue flexible en los tiempos que necesitaba para la facultad o olimpíadas. 
A mis familia, por acompa\~narme y apoyarme en todo lo que pudieran.

\tableofcontents

\mainmatter
\pagestyle{headings}

\chapter{Manga}
Manga is a type of Japanese comic that rose in popularity after World War 2, with works such as \textit{Astro Boy}
in 1952. Today, manga constitutes a great part of Japan industry, influencing television shows, video games, films, music, merchandise and even emojis in social media applications. According to the All Japan Magazine and Book Publisher's and Editor's Association (AJPEA), in 2018 the market totaled 441.4 billion yen (about US\$3.96 billion) while in 2019 it totaled 1.543 trillion yen (about US\$14.12 billion). Digital publishing sales made up 19.9\% of the market in 2019, whereas it made up 16.1\% of the market in 2018. 

Comics can be characterized by its hybrid textual-visual nature \cite{forceville}. Like comic books, manga are composed of four main elements: pictures, words, balloons and panels. Pictures are used to depict objects and figures. Words (including onomatopoeia) indicate character's speech and thoughts. Balloons are used to contain the words and link them to the corresponding character, with different shapes and styles to indicate whether it is speech or thought. Panels are used to structure the narrative, joining together relevant pictures, words and balloons that form a scene and also mark the continuity of time and space by the transitions between. 

\begin{figure}[h!]
\centering
\includegraphics[height=0.7\textheight]{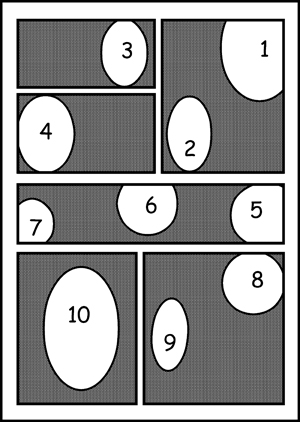}
\caption{Example illustrating the flow to read manga which is from right to left}
\label{mangaflow}
\end{figure}

\begin{figure}[h!]
\centering
\begin{subfigure}[b]{0.73\textwidth}
\includegraphics[width=\textwidth]{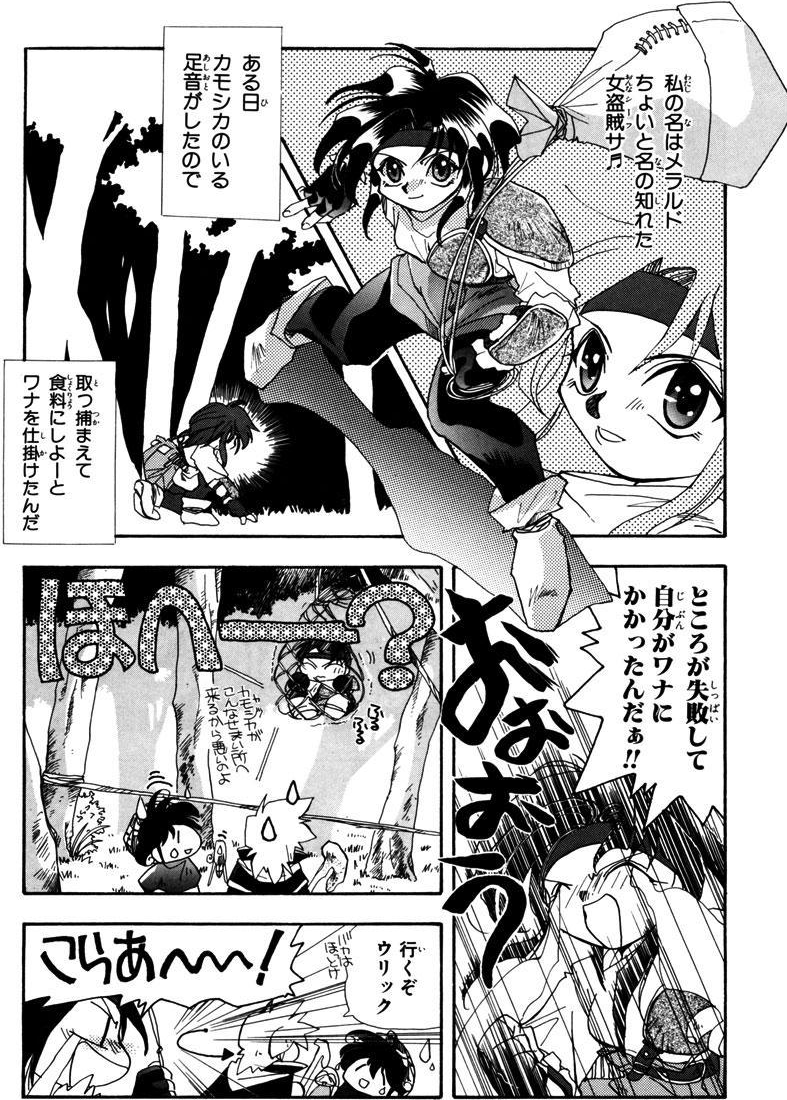}
\caption{}
\label{manga1}
\end{subfigure}
\begin{subfigure}[b]{0.24\textwidth}
\includegraphics[width=\textwidth]{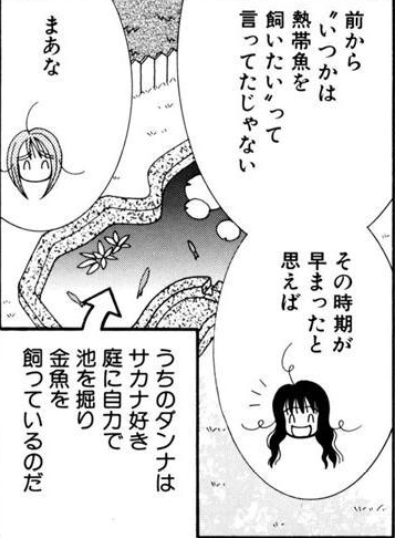} 
\caption{}
\label{manga2}
\includegraphics[width=\textwidth]{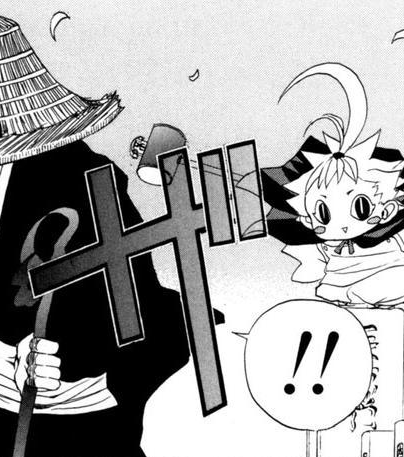}
\caption{}
\label{manga3}
\includegraphics[width=\textwidth]{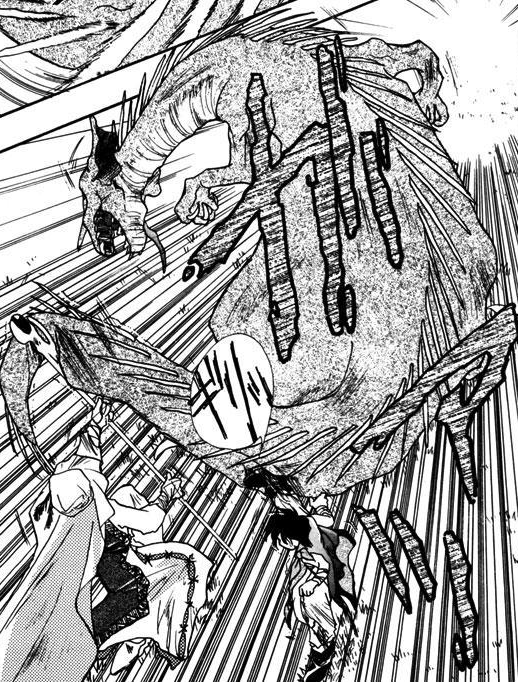}
\caption{}      
\label{manga4}
\end{subfigure}
\caption{
Pictures showing the diversity of text styles in manga. (a) The dialogue balloons could have unconstrained shapes and border styles. The text could have any style and fill pattern, and could be written inside or outside the speech balloons. Note also that the frames could have non-rectangular shapes, and the same character could be in multiple frames. (b) Example of manga extract featuring non-text inside speech bubbles. (c) The same text character can have diverse levels of transparency. (d) Text characters could have a fill pattern similar to the background.
All images were extracted from the \texttt{Manga109} dataset \cite{manga109}\cite{Matsui_2016}\cite{ogawa2018object}:
(a) and (d) ``Revery Earth'' \textcopyright {Miyuki Yama}, (b) ``Everyday Oasakana-chan'' \textcopyright {Kuniki Yuka}, (c) ``Akkera Kanjinchou'' \textcopyright {Kobayashi Yuki} 
}
\label{img:heads}
\end{figure}

They are, however, different from other comics in multiple ways. Unlike American and European comics which tend to be in color, manga is usually in black and white. In manga, the flow of frames and speech go from right to left as seen in Fig. \ref{mangaflow}. While most comics share the same style of art and format, in manga each author tries to add his own style to it. Consequently, there is a huge diversity of text and balloon styles in unconstrained positions (Fig. \ref{img:heads}) compared to comics.

Japanese is a highly complex language, with three different alphabets and thousands of text characters. It also has about 1200 different onomatopoeia, which frequently appear in manga. Japanese language is extremely sophisticated in terms of its ability to express sentiments and emotions though graphic characters. One example is they have three different onomatopoeia to express emptiness, one for the lack of sound, a second one for the lack of motion and a third one to express lack of feeling. This interaction between image and sound can affect translation \cite{Kaindl1999ThumpWP}. Furthermore, characters often look very similar to the art in which they are embedded. These complexities make a text detection method for manga challenging to design.

\begin{figure}[h!]
\includegraphics[width=\textwidth]{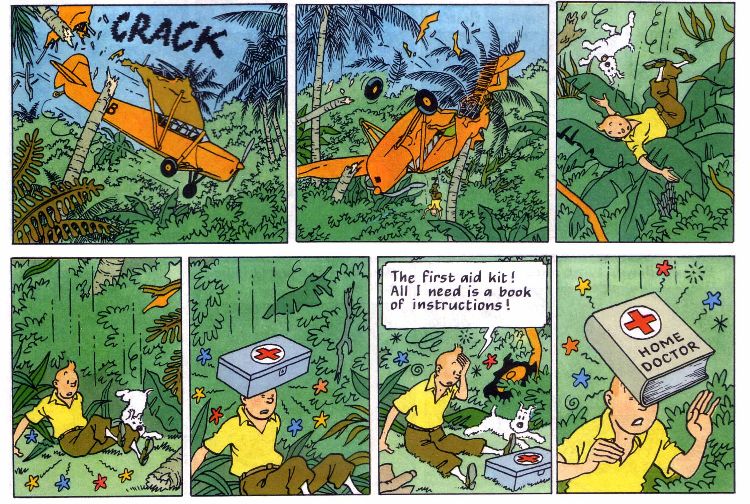}
\caption{Extract of The adventures of Tintin comic}
\label{tintin}
\end{figure}

\begin{figure}[h!]
\centering
\includegraphics[height=0.45\textheight]{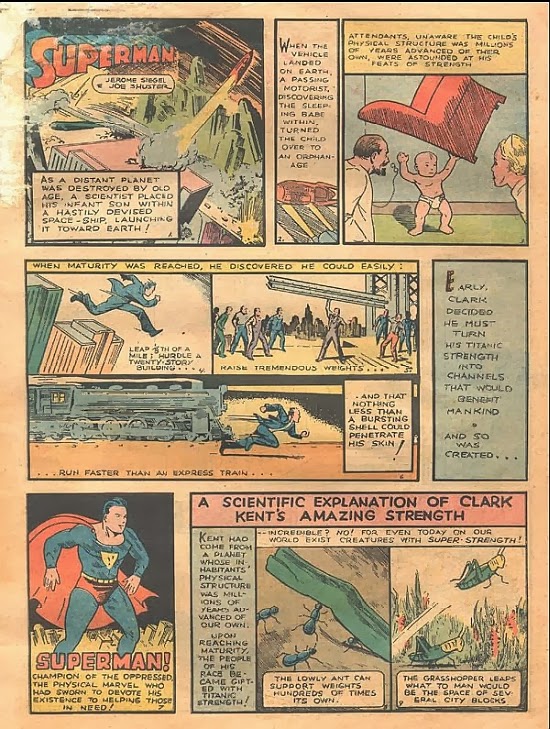}
\caption{Extract of Superman comic}
\label{superman}
\end{figure}

\begin{figure}[h!]
\centering
\includegraphics[height=0.45\textheight]{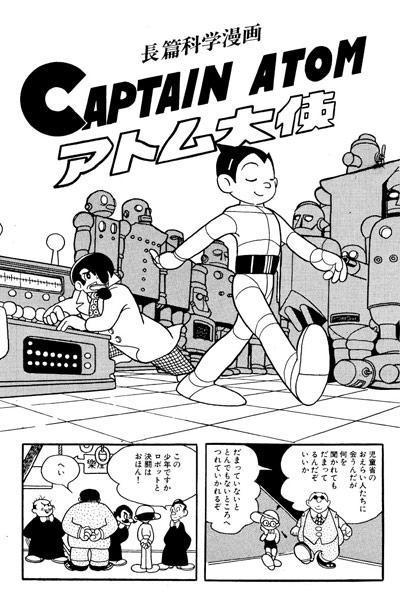}
\caption{Extract of Astro Boy manga}
\label{astroboy}
\end{figure}

\begin{figure}[h]
\includegraphics[width=\textwidth]{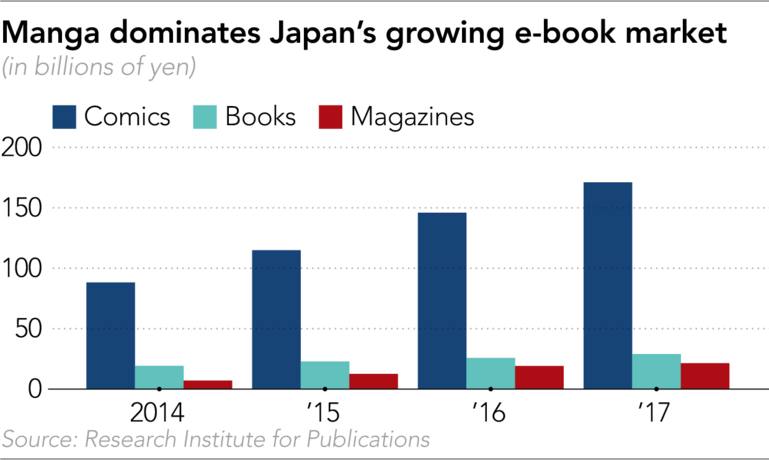}
\caption{Distribution of ebook sales in Japan over the years}
\label{manga_sales}
\end{figure}

United States, France and Japan had the highest influence in the origin of comics and their popularity, each with their different styles integrating diverse elements from their respective cultures. Astroboy (1952, Fig. \ref{astroboy}) identifies Japan culture the same way Superman (1938, Fig. \ref{superman}) does for United States or The adventures of Tintin (1929, Fig. \ref{tintin}) does for France. While United States and France were successful at redistributing their works internationally, manga from Japan was not widely distributed overseas. Few were published abroad and they did not have huge success. It was the internet which spurred its growth in readers worldwide. Even manga ebook sales have been increasing every year in Japan, as shown in Fig. \ref{manga_sales}. While the few who know Japanese could read them, this is definitely not the case for most. The complexity of the Japanese language still hinders its diffusion, even if available digitally. However, many fans work on the arduous process of translating the text in the images to make it available to non Japanese speakers.

This process, known as scanlation, consists of detecting the text, erasing it, inpainting the image, and writing the translated text on the image. As it is an intricate process, the translation is usually done manually in manga, and only the most popular mangas are translated. Automating the translation would lead to solving the linguistic barrier.

In this work, we focus on the first step of the translation process: text detection. 

\chapter{Overview}
Our text detection task is hard, it is still considered an open problem. In order to solve this, we consider the increasingly popular neural networks, briefly discussed in section \ref{neural-networks}. Along with a library in python called fastai \cite{fastai} \cite{fastaip}, we experiment with multiple ideas and provide information on our findings through this journey.

There is an abundance of deep learning related papers, with an increase in every year. On one hand, we benefit from the availability of lots of previous research but on the other hand, this makes it hard to find the specific related research which would be useful for a particular task. Recently, some tools have been made to help searching such as ai index \cite{aiindex}. 

Our problem in particular is hard to search for, as most papers referring to text detection deal with predicting what are the characters written in an image instead of text placement. Furthermore, most of the solutions that involve predicting text placement, do it in the form of bounding boxes or polygons, which are unsuitable for our case. Between both issues, it is hard to search for relevant papers. We had to filter out over 100 text detection related papers in order to find the ones that actually had some relation to our goal. As for Github, there are 2 deep learning projects that deal with detecting text in manga, but both are not peer-reviewed and one does not include the training code. These findings constitute our first contribution in this work, which are highlighted in section \ref{relatedwork}.

Another difficulty we found along this research was that there are very few datasets with pixel accurate labels of where text is placed in an image. This increases the difficulty in finding papers using a pixel level approach. One possibility would be making our own dataset, but this would be a very time consuming process so we decided to first try approaches relying on synthetic generated data. 

Taking ideas from two related papers, we first try to solve the task by text removal, that is to say, taking the image with the text as input and the image without the text already in-painted as output. This is an ambitious goal and we discuss our troubles and findings in section \ref{text-removal}. 

After encountering multiple issues with the results, we decided to change our approach. Instead of text removal, we choose to generate the mask of which pixels are text, usually called text segmentation. This can be found in section \ref{synthetic-segmentation}.

Still unsatisfied with the previous results, we took it further by creating our own dataset \ref{dataset} for this specific task, as there were none available.

Having actual data, it was now also possible to generate more accurate metrics. Which metrics are best for text detection is still an open problem, and few papers have done research about it. We discuss the issues that commonly used metrics have and our choice of metrics to use in section \ref{metrics}.

We employ this dataset and continue with text segmentation, vastly improving previous results and conducting multiple experiments. This is situated in section \ref{model}. Finally, we end with our conclusions in section \ref{conclusions}.

\chapter{Neural Networks}\label{neural-networks}
Neural networks are basically a set of interconnected nodes where each node does some kind of processing to its input and then feeds forward its result to its connections. Using non linear functions as some of those nodes, usually called activation functions, many complex functions can be represented. 

These functions usually have millions of parameters and it is expected that with the right value assignment, many complex tasks can be solved with a performance similar or greater than humans. Finding this set of the right values depends on training, which is done based on examples of inputs and their respective output, thus not needing to code explicitly how to solve the required task.  

With the help of a loss function, which should penalize based on how wrong or right is the output, and an optimizer, which decides how to change the values of the parameters based on this loss, the network is trained and its parameters are increasingly guided to a presumably better set of values suited to the particular task. 

Over recent years, neural networks have grown increasingly popular, both in research and non research communities. One of the main reasons for this is the recent increase in performance, in many tasks the current state of the art involve using a neural network. This is especially true for images, where extracting features is hard and very case specific. 

Current text detection state of the art also involves neural networks, which is why we decided to approach our problem with it.

\section{Frameworks}
The most popular deep learning frameworks are TensorFlow and PyTorch. However, they are quite low level which is why there are many libraries built on top of them like Keras with TensorFlow. Fastai \cite{fastai} \cite{fastaip} is a library built on top of PyTorch which includes many state of the art research into its implementation. This allows easily training models which already come with these features and best practices. Furthermore, it provides enough flexibility to customize all the training process, very useful for research. This is the library we decided to use for this work.

\section{U-Net}
\begin{figure}[h!]
\includegraphics[width=\textwidth]{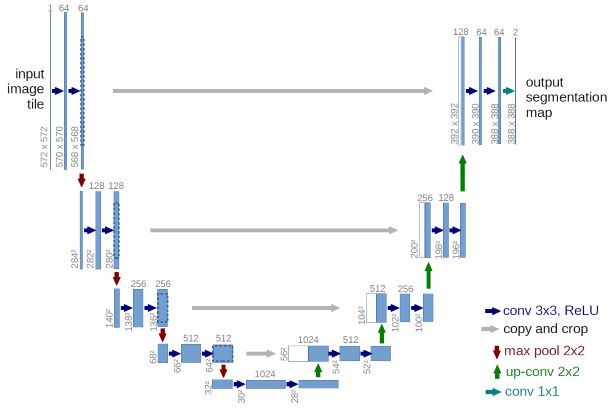}
\caption{Example of U-Net architecture}
\label{unet}
\end{figure}

U-Net \cite{unet} (Fig. \ref{unet}) is a neural network originally designed for medical image segmentation, which became very popular after researchers discovered that it was also achieving state of the art in many other image related tasks. Over the time, many variations have been proposed but the original idea remains the same: in the first part (encoder) the amount of features decrease over further layers while on the second part (decoder) they increase in similar fashion until a similarly sized segmentation map is obtained as output.

The key factor of its success is the cross connections that go from the encoder layers to their respective decoder layers, allowing the model to retain the original information that could have been lost over the down-sampling in order to provide the up-sampling. 

A variation of this architecture, used in this work, is provided by the fastai library  called Dynamic U-Net \cite{fastaiunet} which automatically generates the decoder part based on the encoder.   

\chapter{Related Work}\label{relatedwork}

\hspace{0.4cm} \textbf{Speech balloon detection} Several works have studied speech balloon detection in comics \cite{rigaud,rigaud_2019,liu_extraction,Dubray}. While this could be used to detect speech balloons and then consider its insides as text, the problem is that text in manga is not always inside speech balloons. Furthermore, there are a few cases where not everything inside the balloon is text (Fig. \ref{manga2}).

\textbf{Bounding box detection} Other works in text detection in manga, such as Ogawa \etal \cite{ogawa2018object} and Yanagisawa \etal \cite{Yanagisawa}, have focused on text bounding box detection of multiple objects, including text. Wei-Ta Chu and Chih-Chi Yu have also worked on bounding box detection of text \cite{chu}. 

Without restricting to manga or comics, there are many works every year that keep improving either bounding box or polygon text detection, one of the most recent ones being  Wang \etal \cite{wang2019shape}. However, methods trained with rigid word-level bounding boxes exhibit limitations in representing the text region for unconstrained texts. Recently, Baek \etal proposed a method (CRAFT) \cite{baek2019character} to detect unconstrained text in scene images. By exploring each character and affinity between characters, they generate non-rigid word-level bounding boxes.

\textbf{Pixel-level text segmentation} There are very few works that do pixel-level segmentation of characters, as there are few datasets available with pixel-level ground truth. One of such works is from Bonechi \etal \cite{bonechi2019cocots}. As numerous datasets provide bounding–box level annotations for text detection, the authors obtained pixel-level text masks for scene images from the available bounding–boxes exploiting a weakly supervised algorithm. However, a dataset with annotated bounding boxes should be provided, and the bounding box approach is not suitable for unconstrained text.
Some few works that make pixel text segmentation in manga could be found on GitHub. One is called \enquote{Text Segmentation and Image Inpainting} by \texttt{yu45020} \cite{yu45020} and the other \enquote{SickZil-Machine} by \texttt{KUR-creative} \cite{Sickzil}. Both attempt to generate a text mask in the first step via image segmentation and inpainting with such mask as a second step. In \textit{SickZil-Machine}, the author created pixel-level text masks of the  \texttt{Manga109} dataset, but has not publicly released the labeled dataset. The author neither released the source code of the method but has provided an executable program to run it. In \texttt{yu45020}'s work, the source code has been released, but the dataset used for training is unclear.

\begin{figure}[h]
\centering
\begin{subfigure}[b]{0.45\textwidth}
   \includegraphics[width=\linewidth]{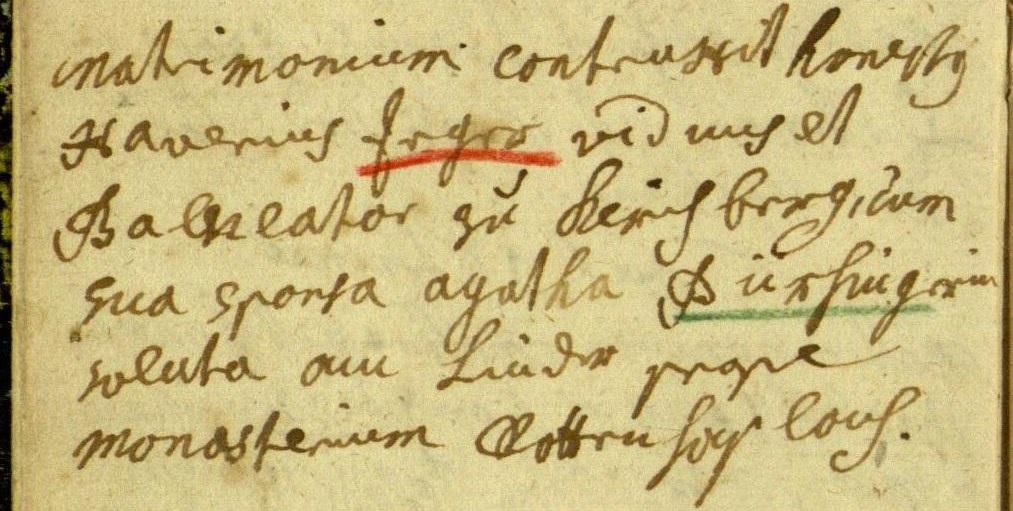}
            \caption{}
\end{subfigure}
\begin{subfigure}[b]{0.45\textwidth}
   \includegraphics[width=\linewidth]{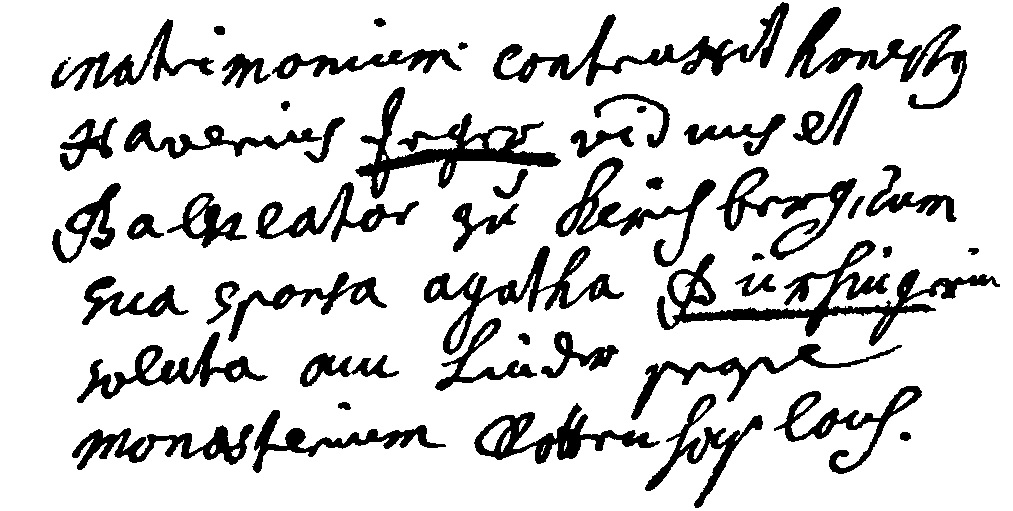}
            \caption{}
\end{subfigure}
\caption{In (a), an image from DIBCO 2018 dataset featuring hand written text. In (b), its ground truth}
\label{fig:dibco}
\end{figure}

We are fully aware that there is a long history of text segmentation and image binarization in the document analysis community related to engineering drawings, maps, letters and more. However, we consider these datasets, where most of the image is text along with a few lines or figures, far more simple than one of manga, which features a lot more context, wide variety of shapes and styles. As an example, in DIBCO 2018 (Document Image Binarization Competition), the dataset is only of 10 images similar to Fig. \ref{fig:dibco}.

\textbf{Text erasers} Some authors have explored pixel-level text erasers for scene images.
Nakamura \etal \cite{nakamura2017scene} is one of the first to address this issue using deep neural networks. Newer works (EnsNet) by Zhang \etal \cite{zhang2018ensnet} and (MTRNet) by Tursun \etal \cite{tursun2019mtrnet} make use of conditional generative adversarial networks.

\chapter{Detecting, removing and inpainting as a single stage}\label{text-removal}
Our first try was to do something similar to EnsNet and MTRNet: erase the text and do the inpainting from the image in a single neural network. In order to train this, both an image with text and an inpainted image without the text is needed. This is hard to come by, be it manga or any kind of image. Thus we proceeded by generating synthetic data.

Danbooru2019 \cite{danbooru2019} is a large scale dataset of anime/manga style images along with tags and other kind of metadata. We downloaded a subset from those and with another text detection software, removed those that already had text. We downloaded several fonts that had Japanese symbols in them, then randomly generated non overlapping rectangles, and in those rectangles randomly generated text. In this way, we had the original image that would be the target of the network and the modified image with text in it as the input, to try to make the network learn to remove Japanese text.

\section{Rectangle generation}
In order to generate non overlapping rectangles, the following approach was used: randomly obtain top left rectangle corner, randomly choose a width and height and if it did not intersect with any of the previous rectangles, add it to our set. If it does intersect, give it a chance to reduce half the width or half the height in order to fit. This process was repeated until either the amount of requested rectangles was reached or a maximum amount of retries was made.

\begin{comment}

\begin{lstlisting}[language=Python]
def generate(width, height, limit):
    rects = []

    for i in range(0, min(limit * 2, 15)):
        x = randint(0, int(width * 0.93))
        y = randint(0, int(height * 0.9))
        
        if random_sample() < 0.8:
            w = randint(7, 15)
            h = randint(10, 35)
        else:
            w = randint(15, 100)
            h = randint(10, 50)
        
        w = min(int(w * width / 100), width)
        h = min(int(h * height / 100), height)
        r = Rectangle(x, y, w, h)
        add = True
        
        for rect in rects:
            if rect.intersects(r) and random_sample() < 0.5:
                r = Rectangle(x, y, int(r.width / 2), r.height)
                if rect.intersects(r) and random_sample() < 0.5:
                    r = Rectangle(x, y, r.width, int(r.height / 2))
            if rect.intersects(r):
                add = False
                break
        
        if add:
            rects.append(r)
            if len(rects) == limit:
                break
    return rects    
\end{lstlisting}
\end{comment}

\section{Text generation}
To generate a random text of $n$ characters, we simply randomly choose characters $n$ times from the unicode code point ranges that include japanese characters, along with some special symbols and english letters.

Given a rectangle width and height, we need to make sure the text will fit inside the rectangle. In order to do this, we must make sure that if drawing the text with the given font would overflow in width, we either send the rest to a new line and continue processing if there is still enough height or we just cut the text there. This problem is know as text wrapping.

Many ways of doing this can be found online, but most were very inefficient or handled different kind of wrapping such as no more than x characters per row regardless of pixel width, taking care not to split words. 

An algorithm that provides the exact solution for pixels is:
\begin{lstlisting}[language=Python]
def text_wrap(text, font, max_width, max_height):
    lines = []
    i, j, hei = 0, 0, 0
    while j <= len(text):
        w = font.getsize(text[i:j + 1])[0]
        if w > max_width or j == len(text):
            hei += font.getsize(text[i:j])[1]
            if hei <= max_height and j > i:
                lines.append(text[i:j])
                i = j
                if j == len(text):
                    break
            else:
                break
        else:
            j += 1 
    return lines
\end{lstlisting}

While this makes sure text always stays within the rectangle, it is very slow. The calls to getsize are the ones that take most time, so our goal is trying to use as few as possible. After trying out several options, we ended up with the following version which is 5 to 10 times faster in most cases:

\begin{lstlisting}[language=Python]
    def text_wrap(text, font, max_width, max_height):
        estimate = (max_width // font.getsize('a')[0])
        lines = []
        i, j, hei = 0, 0, 0
        while i < len(text):
            i = j
            j = min(len(text), i + estimate)
            width = font.getsize(text[i:j])[0]
            while j < len(text) and width <= max_width:
                width += font.getsize(text[j])[0]
                j += 1
            while width > max_width and j > i:
                j -= 1
                width -= font.getsize(text[j])[0]
            hei += font.getsize(text[i:j])[1]
            if hei > max_height:
                break     
            if len(text[i:j]): 
                lines.append(text[i:j])  
        return lines
\end{lstlisting}

\section{Fonts}
While it is easy to download many fonts, its not as easy to know if a font supports a certain character. In all the tools or code we found, it was always wrong. There are several font formats, but in most there is a character that is defined as the default when a font can't draw a character as it does not support it. Most seem to use the same character for this (0x1d). While we randomly choose characters out of 21275, it may happen that the font we are using only supports a few hundred. 

\begin{figure}[h!]
\center
\includegraphics[width=0.3\textwidth]{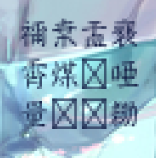}
\caption{Drawing text over an image, but 3 of the characters are not supported by the font}
\label{missingchar}
\end{figure}

This leads to a lot of characters being drawn as the default missing one on the images as seen in Fig. \ref{missingchar}. This is a big problem as we are wasting a lot of learning potential for the network. We believe that the online tools and code are probably working fine, but the font instead of properly having all the unsupported characters as missing, it actually defines the mapping to the missing character. In the end, we were able to design a method to discard these characters, although it may be discarding more than necessary.

\section{Textify}
We called textify to our method of adding text to an image. While this changed many times over time, we show the pseudocode of the final version:
\begin{lstlisting}[basicstyle=\small]
padding = randint(4, 10)
with 50% chance:
    generate a single rectangle that covers at least 66% of the image
else:
    generate between 7 and 15 rectangles with previous method

for each rectangle:
    get random text size, mostly regular sizes.
    estimate how many characters will fit in the rectangle
    pick a random font, weighted by amount of characters it supports
    generate random text, with length = 50% to 100% of the estimation
    split text into lines as defined by text_wrap
    randomly choose the color text will have, mostly black
    randomly choose border color, mostly white
    with 30% chance, decide if text will be rotated in random angle
    with 10% chance, decide if the rectangle will also be drawn
    with 10% chance, decide if the rectangle will be semi transparent
    with 5% chance, decide if border will be added
    with 50% chance, change the rectangle to an ellipse
    Finally draw the text in the image, applying all the decisions

with 20% chance, convert image to black and white
\end{lstlisting}

All these random transformations attempt to make the synthetic data cover as many cases as possible, to force the network focus on text.

\section{Metrics}
Standard L1 or L2 sum over the pixels are not a good measure to compare results in many image to image tasks, such as inpainting. This is still an open problem and many ways to compare image similarity are designed every year. From those, we chose the most popular: SSIM (Structural Similarity Index) \cite{ssim} and PSNR (Peak Signal-to-noise Ratio).
        
\section{Loss function}
As L1 or L2 are not very useful as metric, they are also not very useful as loss function. Instead we use a feature loss function, which considers features obtained from vgg16 model, a similar approach to \cite{johnson2016perceptual}: 

\begin{equation}
loss(i, t) = L_{1}(i, t) + \sum_{j=0}^{j=n}L_{1}(f_{j}(i), f_{j}(t))*w_{j} + L_{1}(gm(f_{j}(i)), gm(f_{j}(t)))*w_{j}*5e3
\end{equation}
where $i$ is the input image, $t$ is the target image, $f_{j}$ is the $j$th vgg16 model features of the $n$ selected layers, $w_{j}$ is a predefined weight and $gm$ is the gram matrix.

\section{Training}
Initially, we used 5000 images cropped to 64x64 to train the U-net with a resnet18 encoder. Normalizing dataset and setting a sigmoid as layer to force output to be in the range -1 to 1 helped get better results. Parameters like self attention or blur did not seem to have any noticeable effect.

Trying resnet34 encoder, didn't get noticeable improvements either. With resnet101 it did, but took much longer to train. Using variations of U-net, U-net wide didn't improve while U-net deep did, but took much more memory and took longer to train.

In the end, we decided to keep the U-net with the resnet18 encoder as it took much less to train, leading to faster testing different settings, and the results weren't much worse. 

\section{Problems}
\begin{figure}[h!]
\center
\includegraphics[height=0.2\textheight]{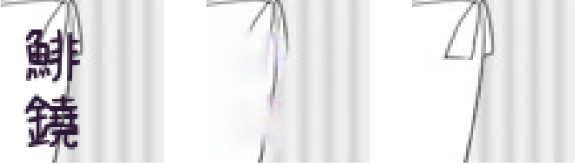}
\caption{Input, output and ground truth patches in each column. Edges are lost, but more importantly even in an “easy” case of mostly white and black text, patched zone remains very blurry}
\label{artifacts1}
\end{figure}

\begin{figure}[h!]
\center
\includegraphics[height=0.2\textheight]{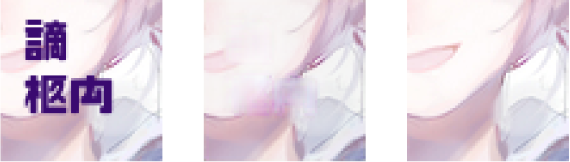}
\caption{Input, output and ground truth patches in each column. Blurring is even worse in color patches}
\label{artifacts2}
\end{figure}

\begin{figure}[h!]
\center
\includegraphics[height=0.2\textheight]{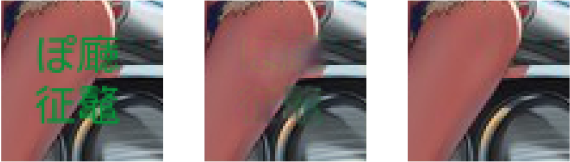}
\caption{Input, output and ground truth patches in each column. Text not completely removed}
\label{artifacts3}
\end{figure}

\begin{figure}[h!]
\center
\includegraphics[height=0.2\textheight]{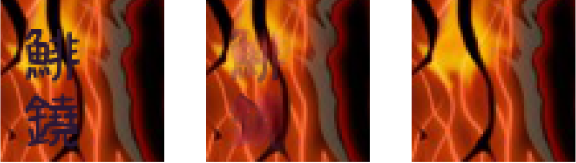}
\caption{Input, output and ground truth patches in each column. Even black text in color patches is not completely removed}
\label{artifacts4}
\end{figure}

As for the PSNR metric, the higher the better. Most experiments lead to 28-29 score, and it was difficult to observe any difference. Anything over 29, however, was noticeable better. Even in those cases, predictions still suffered from multiple artifacts such as blurring (Figs. \ref{artifacts1} and \ref{artifacts2}) or text not completely removed (Figs. \ref{artifacts3} and \ref{artifacts4}).

\subsubsection{Hypothesis}

To test if the model was not capable of reconstructing image, adding text was removed to try training the image identity. Initial efforts seemed to reach similar metrics as with text, but with reaching up to 32 PSNR. After trying other parameters, it was able to learn it completely: 49 PSNR and 99.9 SSIM. 

Given that the model was able to learn the identity and that networks with more parameters like resnet101 did not help solve these issues either, it didn’t seem to be a problem of model not having the capacity to learn it.

Two likely suspects to explain it were that the images were too small and 64x64 was not enough for the model to learn finer details or that the dataset had too few images. It seemed unlikely to be a problem of not seeing enough examples of text, because they were randomly placed and randomly generated, giving a very high possible amount of examples. Even if it was only 5000 images, by using patches of 64x64 of the original 512x512 image, each image could also provide different patches, changing even more the amount of possible examples.

\subsubsection{Testing}

The most likely suspect seemed to be the patches being too small, as during a small text with 64x128, better results were already obtained. To try this out, an experiment was done training model in different stages, progressively increasing the size. For this, 2 parameters were set for each stage: the minimum size of the patch and the maximum, making variable size possible under different batches. These values applied for both height and width, making rectangle patches also possible.

The configuration of these parameters were: start with 64x64 patches, then 64x128, then 128x128, then 128x256 and finally 256x256. Instead of the 5000 images, the full dataset (25000 images) was used.

\begin{figure}[h!]
\center
\includegraphics[height=0.5\textheight]{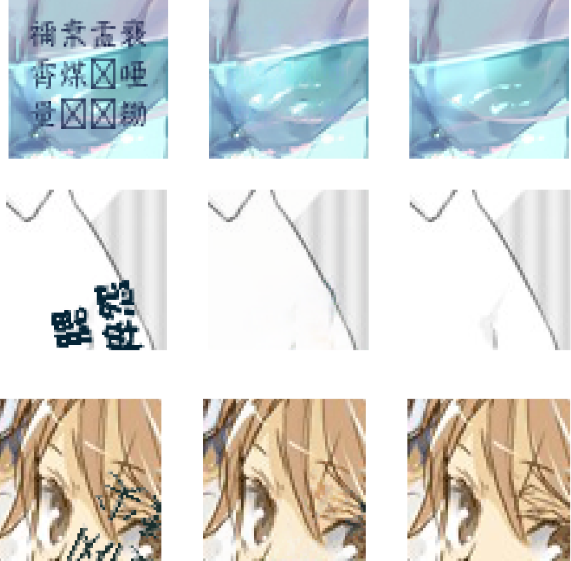}
\caption{Results after first stage of training the 64x64 patches}
\label{improvements1}
\end{figure}

The first noticeable difference was that with just the first stage, the results were remarkably better: not only did it reach better metrics (31.5 PSNR and 0.969 SSIM), but also the erasing and inpainting improved remarkably as seen in Fig. \ref{improvements1}. Some colors were still off and a bit blurry but it was much better than before and even the edge was reconstructed. Given that the same parameters were used, it seemed to be a case of just needing more training data.

\begin{figure}[h!]
\center
\includegraphics[height=0.35\textheight]{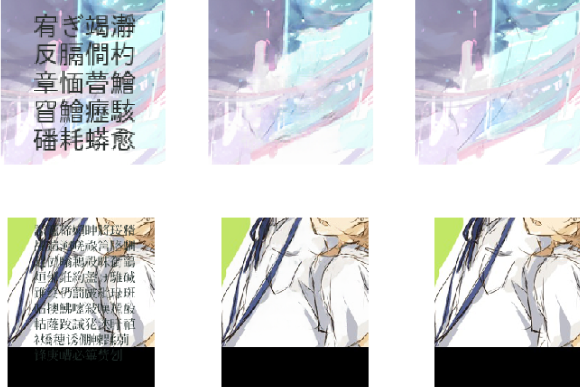}
\caption{Input, output and ground truth patches in each column. Results after 128x256 stage}
\label{improvements2}
\end{figure}

With the progressive resizing, the metrics worsened a bit but the results were still very good, after the (128, 256) stage it had 30.83 PSNR and 0.960 SSIM. This means that the bigger the image, the more likely to have lower PSNR. As seen in Fig. \ref{improvements2}, blurring is much less noticeable and text is completely removed, even in color examples.

By the (256, 256) stage, as the image size was bigger, instead of just putting a single bunch of text centered in the patch, several portions of text were placed over the image, thus making the amount of text lower but including different examples (fonts, color, font size) in a single image.

\begin{figure}[h!]
\center
\includegraphics[height=0.35\textheight]{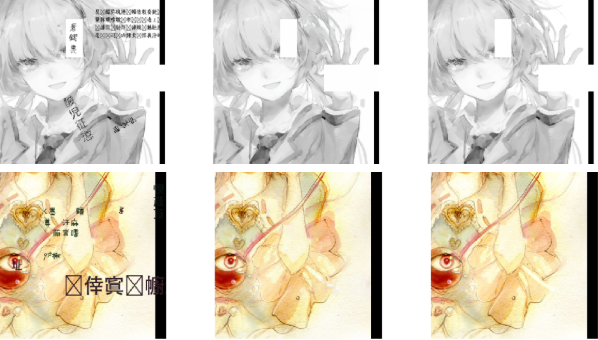}
\caption{Input, output and ground truth patches in each column. Results after 256x256 stage}
\label{improvements3}
\end{figure}

First epoch of this stage already had 37.05 PSNR and 0.991 SSIM, and by the final epoch it reached 38.56 PSNR and 0.993 SSIM. This seems to be a great improvement, but the change is mostly caused by changing the amount of text in the image. With fewer pixels with text, less pixels need to be inpainted so those metrics now give perfect results for a higher percentage of pixels. At first glance, results seem perfect as seen in Fig. \ref{improvements3}.

\begin{figure}[h!]
\center
\includegraphics[height=0.45\textheight]{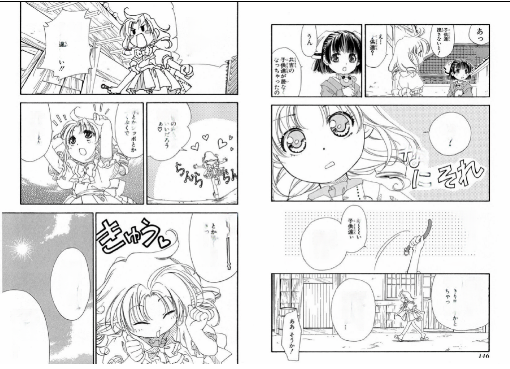}
\caption{Prediction of 256x256 stage}
\label{256x256_1}
\end{figure}

\begin{figure}[h!]
\center
\includegraphics[height=0.45\textheight]{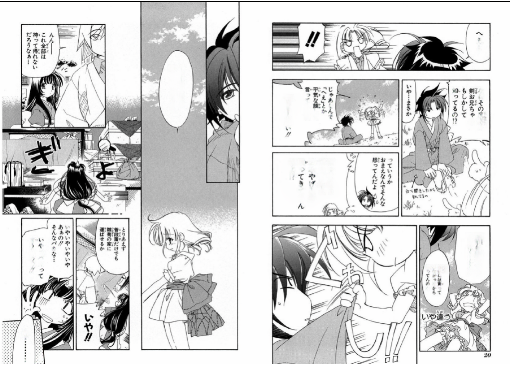}
\caption{Prediction of 256x256 stage}
\label{256x256_2}
\end{figure}

\begin{figure}[h!]
\center
\includegraphics[height=0.45\textheight]{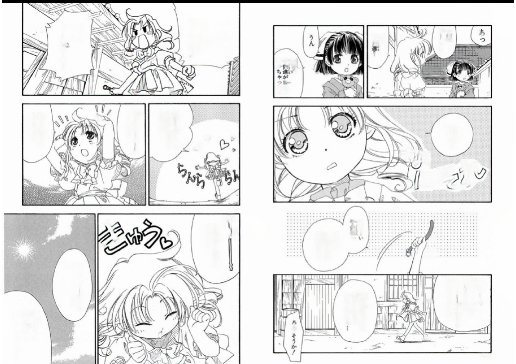}
\caption{Prediction of 128x256 stage}
\label{128x256_1}
\end{figure}

\begin{figure}[h!]
\center
\includegraphics[height=0.45\textheight]{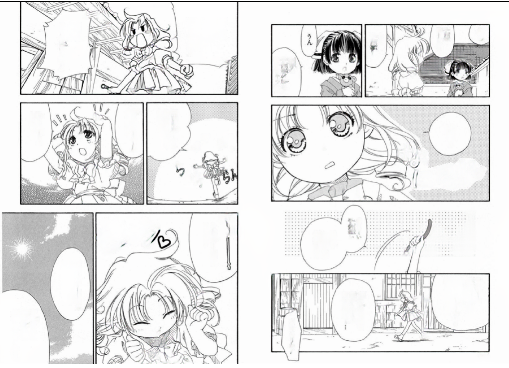}
\caption{Prediction of 128x128 stage}
\label{128x128_1}
\end{figure}

\begin{figure}[h!]
\center
\includegraphics[height=0.45\textheight]{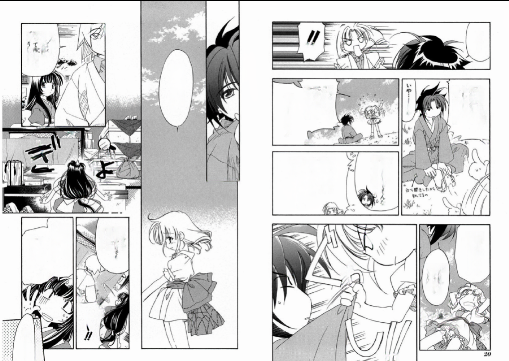}
\caption{Prediction of 128x128 stage}
\label{128x128_2}
\end{figure}

\begin{figure}[h!]
\center
\includegraphics[height=0.45\textheight]{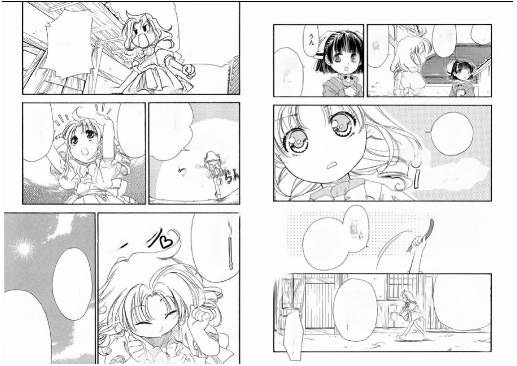}
\caption{Prediction of 64x128 stage}
\label{64x128_1}
\end{figure}

\begin{figure}[h!]
\center
\includegraphics[height=0.45\textheight]{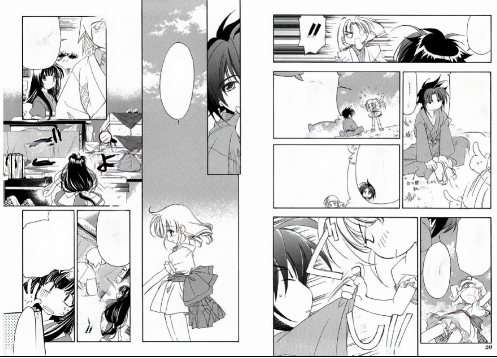}
\caption{Prediction of 64x128 stage}
\label{64x128_2}
\end{figure}

\begin{figure}[h!]
\center
\includegraphics[height=0.45\textheight]{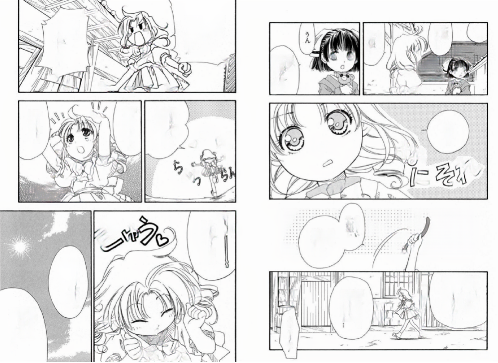}
\caption{Prediction of 64x64 stage}
\label{64x64_1}
\end{figure}

\begin{figure}[h!]
\center
\includegraphics[height=0.45\textheight]{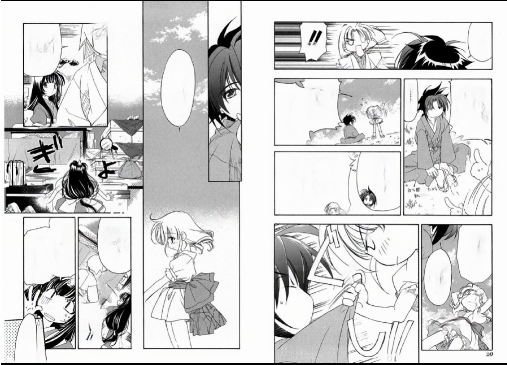}
\caption{Prediction of 64x64 stage}
\label{64x64_2}
\end{figure}

However, when predicting over our actual images from manga, we noticed that the performance was actually much worse in this stage than earlier stages. Furthermore, the predictions seem to be the best at the second stage of 64x128 (Figs. \ref{256x256_1}, \ref{256x256_2}, \ref{128x256_1}, \ref{128x256_1}, \ref{128x128_1}, \ref{128x128_2}, \ref{64x128_1}, \ref{64x128_2}, \ref{64x64_1}, and \ref{64x64_2}). A possible explanation is that it ended up over-fitting to the style of synthetic text we generated, which is different from the one in real manga.

\begin{figure}[h!]
\center
\includegraphics[height=0.45\textheight]{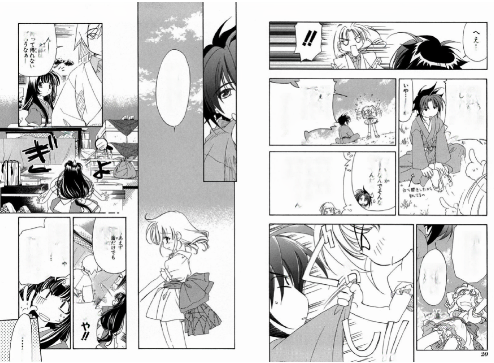}
\caption{Prediction of 128x256 stage}
\label{128x256_2}
\end{figure}

\clearpage

\FloatBarrier
\subsubsection{Resizing}
\begin{figure}[h!]
\center
\includegraphics[height=0.35\textheight]{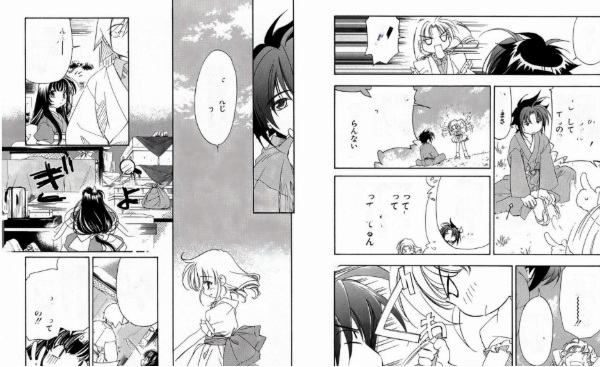}
\caption{Prediction of 64x128 stage with image resized to 1170x1654}
\label{resize_1170x1624}
\end{figure}

\begin{figure}[h!]
\center
\includegraphics[height=0.35\textheight]{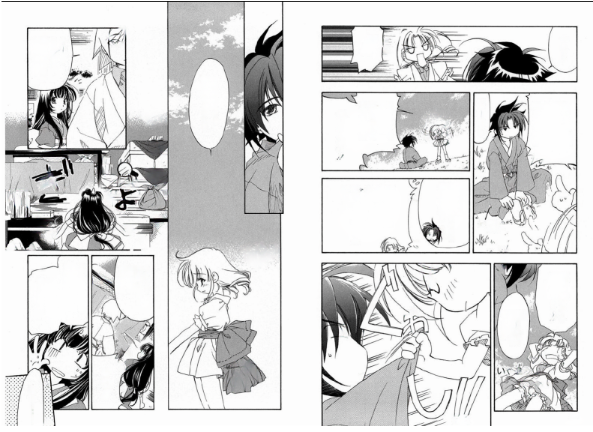}
\caption{Prediction of 64x128 stage with image resized to 1600x800}
\label{resize_1600x800}
\end{figure}

Another important factor that changed a lot the predictions over the manga images was the resizing of the image before the prediction (Figs. \ref{resize_1170x1624} and \ref{resize_1600x800}). Instead of just resizing by stretching, using black padding on the relevant dimension worked better. 

\clearpage

\FloatBarrier
\subsubsection{Limitations}
\begin{figure}[h!]
\center
\includegraphics[height=0.45\textheight]{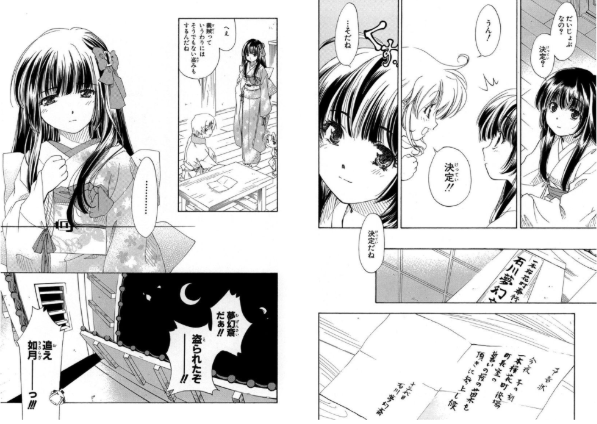}
\caption{Original image}
\label{issues1}
\end{figure}

\begin{figure}[h!]
\center
\includegraphics[height=0.45\textheight]{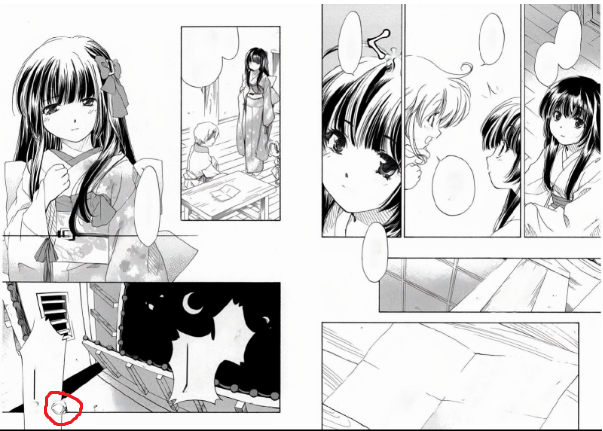}
\caption{Prediction. Circled in red, an example of non text that suffered from removal}
\label{issues2}
\end{figure}

\begin{figure}[h!]
\center
\includegraphics[height=0.45\textheight]{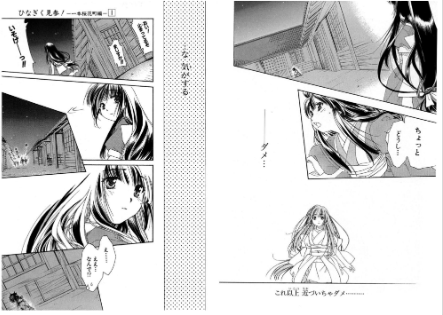}
\caption{Original image}
\label{issues3}
\end{figure}

\begin{figure}[h!]
\center
\includegraphics[height=0.45\textheight]{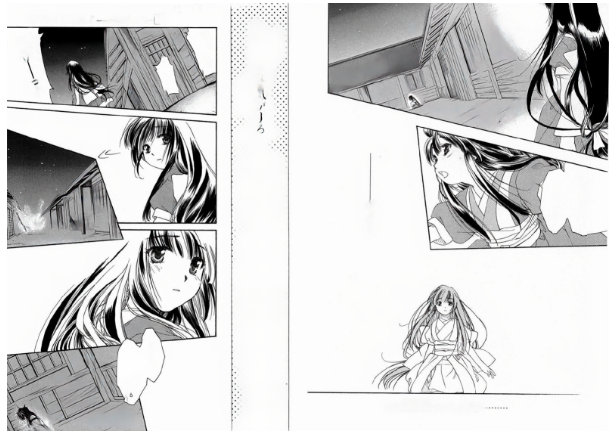}
\caption{Prediction. Many small circles were removed, even when they were not text}
\label{issues4}
\end{figure}

After many tries and improvements, good results were obtained but they still several issues. Firstly, as seen in Figs. \ref{issues1} and \ref{issues2}, although it did a great job of erasing text from speech bubbles and even worked on the letter which not only has rotated text but also perspective, it erases more than necessary (has false positives) such as the face of the dialogue. Secondly, this issue is even more noticeable with small circles, as it tends to erase them as seen in Figs. \ref{issues3} and \ref{issues4}.

Many variations were tried: changing learning rates, number of epochs, size of the patches, the sigmoid range, the amount of fonts used. However, these issues still persisted. This lead us to change our approach.

\chapter{Segmentation on synthetic images}\label{synthetic-segmentation}
Even if we had good results with some images, they were still not perfect. This means that in order for it to be useful, someone would need to fix the mistakes. This not only meant erasing and inpaiting manually the text that was not removed but also recovering the parts that were mistakenly removed. This seemed too troublesome to do, and wouldn't be much different than processing the whole image manually.

In order to let users fix the mistakes more easily, an additional stage would need to be introduced. An alternative approach then is, instead of doing the removal and inpainting of text as a single task, first detect the text and then inpaint it, with 2 separate networks. As several inpainting works already exist, we decided to focus on the first part: text detection.

On one side, many works in text detection in comics have taken a balloon detection approach. However, in manga, the text and balloons are also part of the artwork. Thus, balloons could have a multiplicity of shapes and styles. Besides, the text can be outside the dialogue balloons (Figs. \ref{manga1}, \ref{manga3}, \ref{manga4}), or inside the balloon there could be non-text contents (Fig. \ref{manga2}), making a balloon detection approach unsuitable for this task. On the other side, most previous works in text detection have taken a box detection approach. However, manga contains texts that are deformed, extremely large, or are drawn on the cartoon characters, which are hard to identify with a single bounding box (Figs. \ref{manga1}, \ref{manga3}, \ref{manga4}).

Thus, we decide to make text segmentation at a pixel level, identifying pixels as either text or background.

\section{Danbooru2019 results}
\begin{figure}[h!]
\center
\includegraphics[height=0.425\textheight]{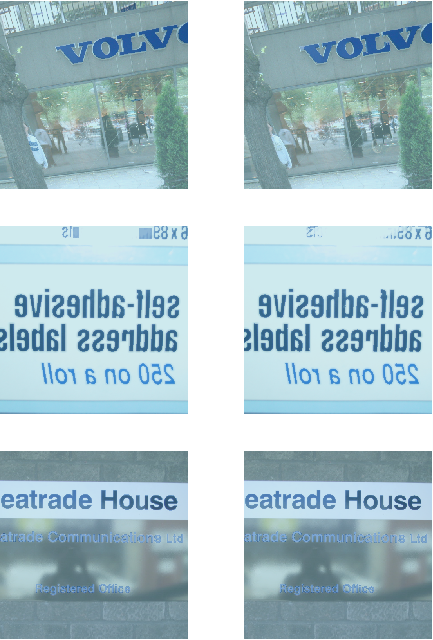}
\caption{Results of icdar2013 training. First column represents ground thruth, while second column is the prediction}
\label{icdar2013}
\end{figure}

In order to quickly test if segmentation was a viable approach, we first tried training on an existent dataset for text segmentation called icdar13 \cite{icdar2013}. As seen in Fig. \ref{icdar2013} results were pretty accurate, so we decided segmentation was indeed a feasible approach.

\begin{figure}[h!]
\center
\includegraphics[height=0.425\textheight]{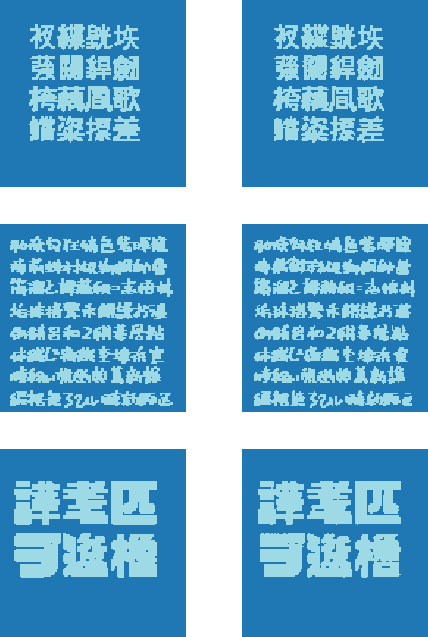}
\caption{Results of Danbooru2019 training over 128x128 patches. First column represents ground thruth, while second column is the prediction}
\label{danbooru1}
\end{figure}

\begin{figure}[h!]
\center
\includegraphics[height=0.35\textheight]{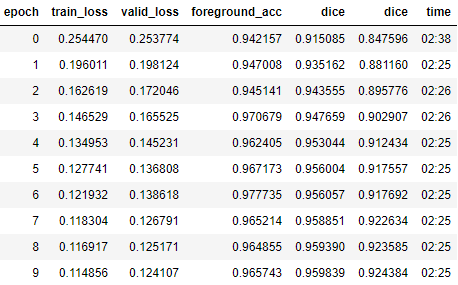}
\caption{Table of metrics and loss over 10 epochs of training Danbooru2019 dataset over 128x128 patches. Second dice metric is with iou (intersection over union) parameter in true}
\label{danbooru1-train}
\end{figure}

We then modified our code to treat the added text as the target instead of the image without it represented as a binary map. The loss function was changed to dice loss and the results over the synthetic data generated with Danbooru2019 were good as seen in Figs. \ref{danbooru1} and \ref{danbooru1-train}.

\section{Manga results}

\begin{figure}[h!]
\center
\includegraphics[height=0.425\textheight]{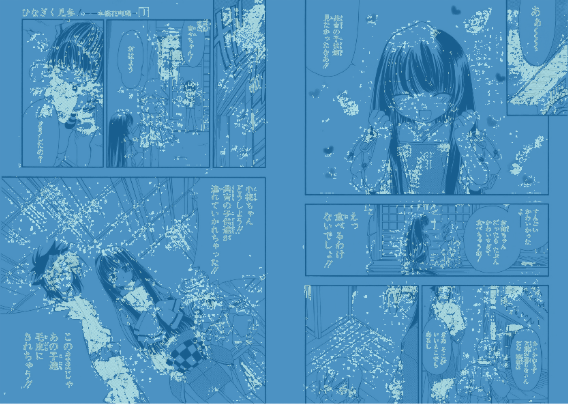}
\caption{Result of Danbooru2019 training over 128x128 patches over manga image with 0.5 as threshold}
\label{danbooru1e1}
\end{figure}

\begin{figure}[h!]
\center
\includegraphics[height=0.425\textheight]{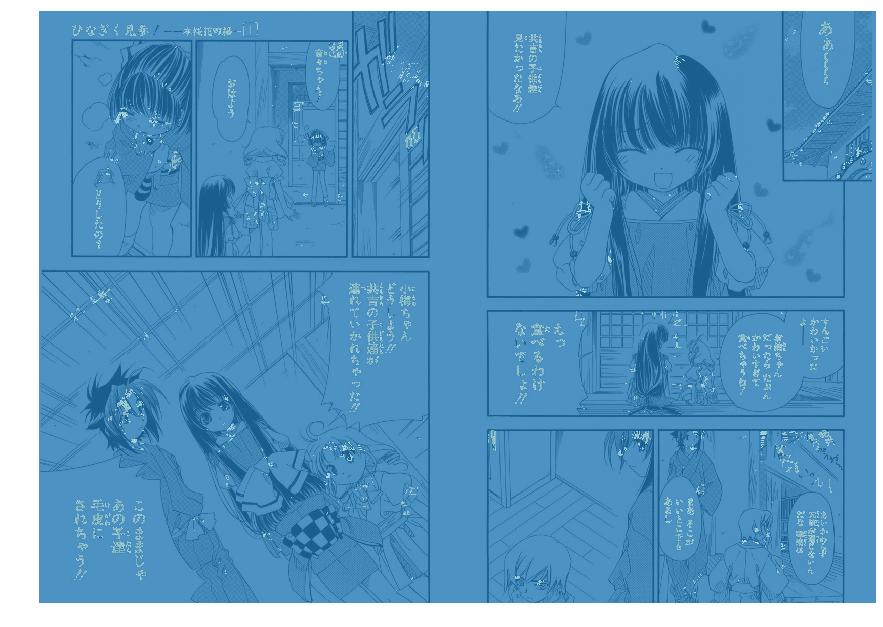}
\caption{Result of Danbooru2019 training over 128x128 patches over manga image with 0.95 as threshold}
\label{danbooru1e2}
\end{figure}

However, when testing with the actual manga images, results were much worse: while most of the text was accurately covered, it had too many false positives. This can be observed in Fig. \ref{danbooru1e1}. Changing the threshold to consider a pixel as text to 0.95 instead of the 0.5 default however, greatly improved the result as seen in Fig. \ref{danbooru1e2}

\clearpage

\begin{figure}[h!]
\center
\includegraphics[height=0.425\textheight]{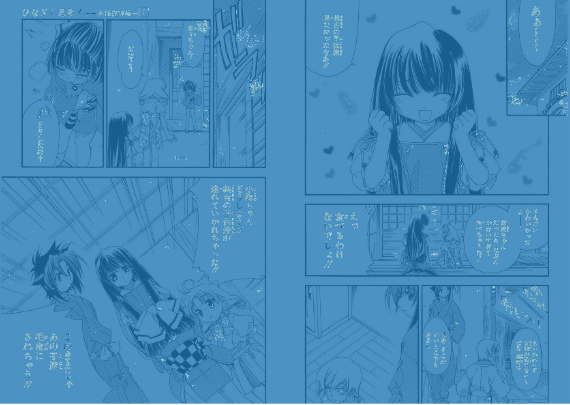}
\caption{Result of Danbooru2019 refinement with 0.95 as threshold}
\label{danbooru2e1}
\end{figure}

\begin{figure}[h!]
\center
\includegraphics[height=0.425\textheight]{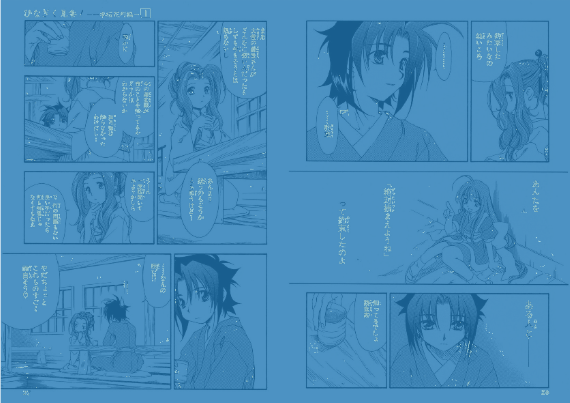}
\caption{Result of Danbooru2019 refinement with 0.95 as threshold}
\label{danbooru2e2}
\end{figure}

\begin{figure}[h!]
\center
\includegraphics[height=0.425\textheight]{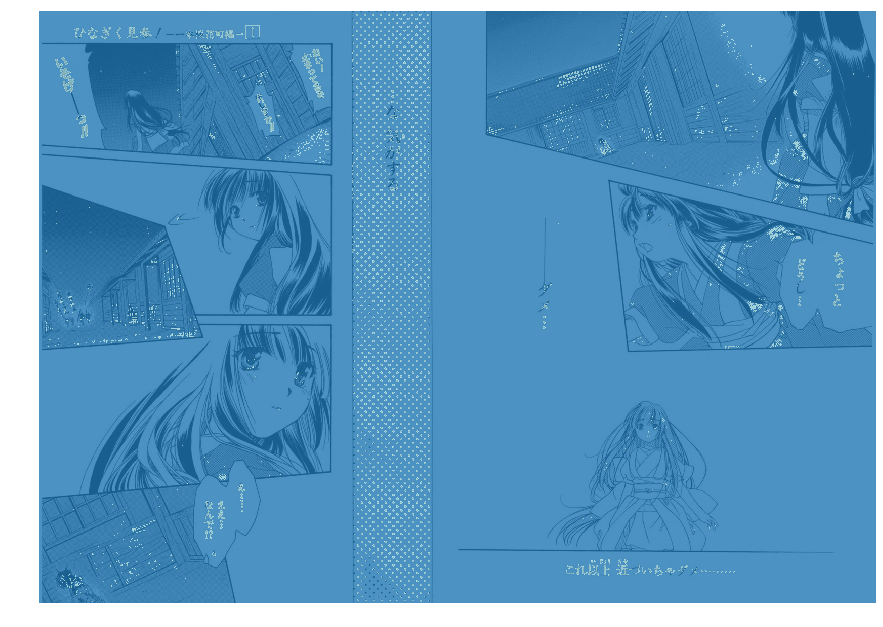}
\caption{Result of Danbooru2019 refinement with 0.95 as threshold}
\label{danbooru2e3}
\end{figure}

\begin{figure}[h!]
\center
\includegraphics[height=0.425\textheight]{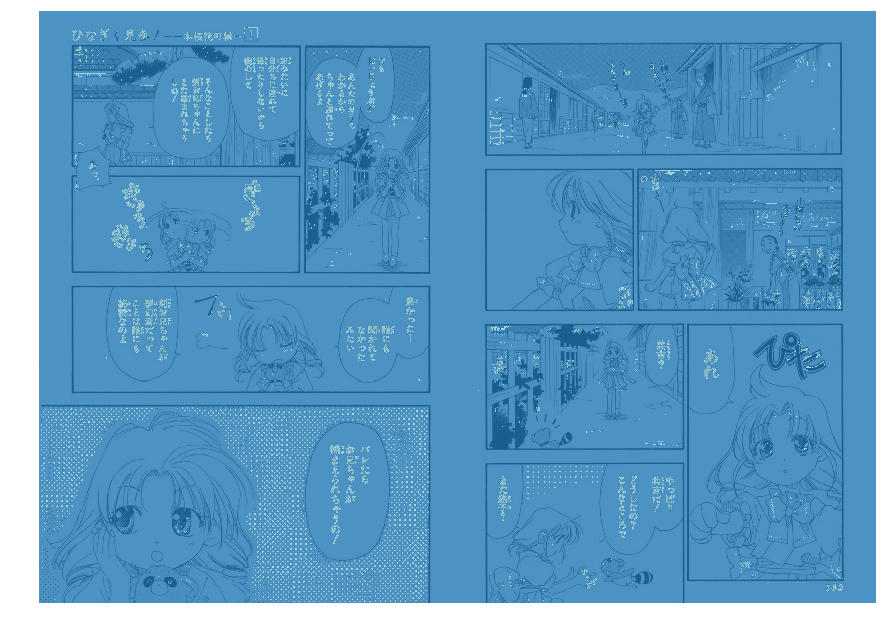}
\caption{Result of Danbooru2019 refinement with 0.95 as threshold}
\label{danbooru2e4}
\end{figure}

\begin{figure}[h!]
\center
\includegraphics[height=0.425\textheight]{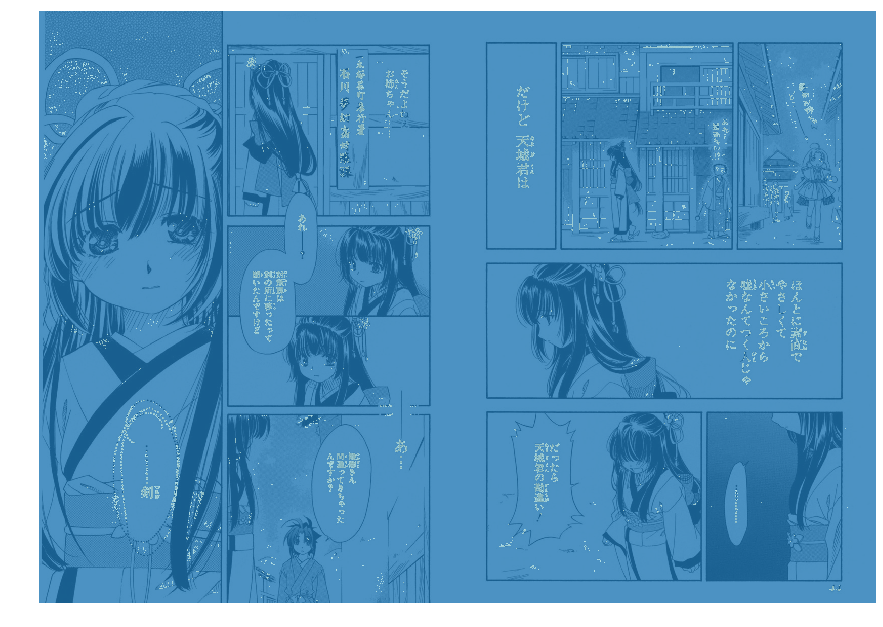}
\caption{Result of Danbooru2019 refinement with 0.95 as threshold}
\label{danbooru2e5}
\end{figure}

Further refinement of hyper-parameters such as max learning rate, number of epochs, loss functions (dice, binary cross entropy, focal loss) and self-attention were made and results were slightly improved as seen in Figs. \ref{danbooru2e1}, \ref{danbooru2e2}, \ref{danbooru2e3}, \ref{danbooru2e4} and \ref{danbooru2e5}.

\clearpage

\begin{figure}[h!]
\center
\includegraphics[height=0.3\textheight]{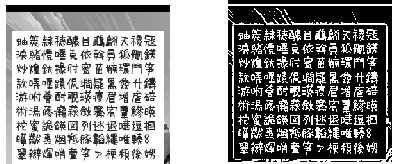}
\caption{Image binarization example}
\label{binarization}
\end{figure}

Training with bigger patches such as 256x256 or 512x512 lead to worse results. While there were less false positives of little dots, less characters were correctly segmented. Another test was applying some pre-processing (image binarization) to the input image in order to make it easier to learn as seen in Fig. \ref{binarization} but it also lead to worse results, which means the network takes advantage of more information (gray scale values).

Other ideas tried were: adding text over manga patches instead of Danbooru2019, deep unet and wide unet variations, filtering the Danbooru2019 images to remove the ones with text, mish activation funcion. None improved the results, so we decided to stop trying to improve the network. 

\section{Post-processing}
What could still improve however, was the post-processing. We tried multiple threshold methods from cv2 and scikit image libraries, along with some noise cleaning methods. The goal was to remove as much as possible the small false positive dots without interfering with the correct text results.  

The algorithm can be seen in listing \ref{noise}. Basically, we assume that contours with high area ($\geq$ 100) are correct predictions. For every contour with less than 100 area, we need to decide if it should be discarded or not. To do that, contours retrieved by cv2 in close order to it are checked to see if they are correct and are close to the current contour. If that is true, the current contour with small area is considered to be correct. The algorithm keeps running until no more small contours are considered to be good.

\begin{figure}[h!]
\center
\includegraphics[height=0.45\textheight]{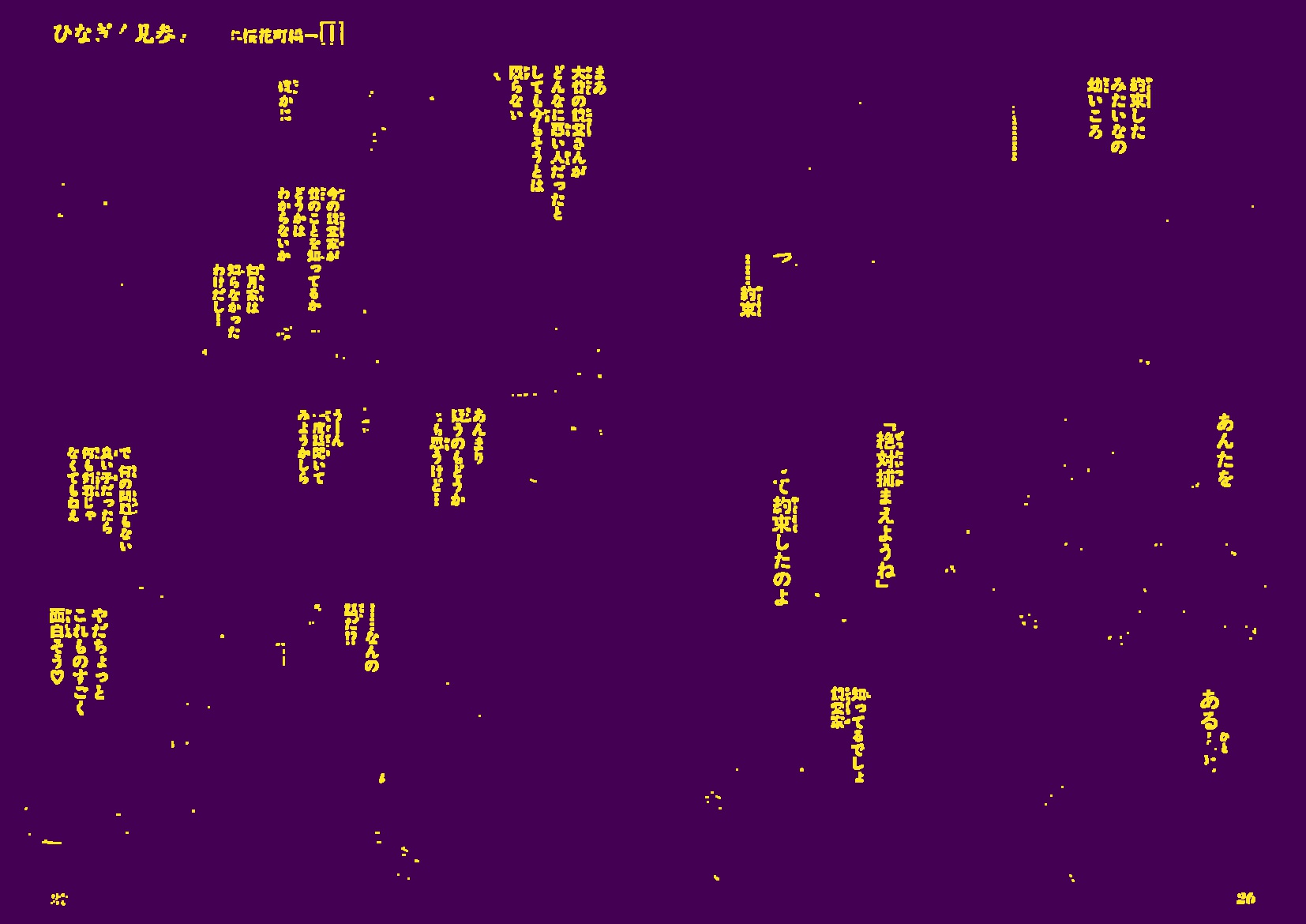}
\caption{Prediction before applying noise removal}
\label{post-processing-1}
\end{figure}

\begin{figure}[h!]
\center
\includegraphics[height=0.45\textheight]{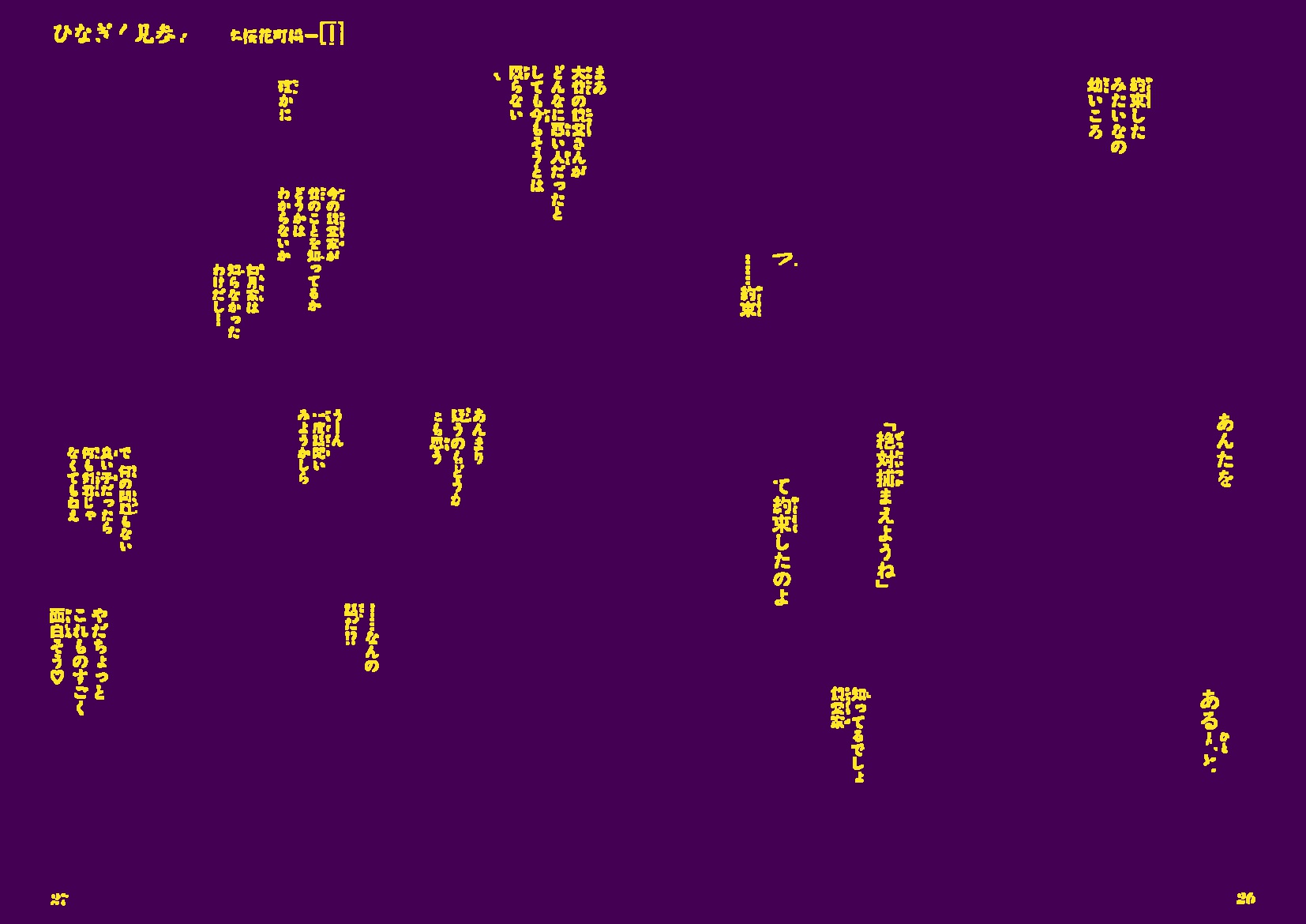}
\caption{Prediction after applying noise removal}
\label{post-processing-2}
\end{figure}

\begin{figure}[h!]
\center
\includegraphics[height=0.45\textheight]{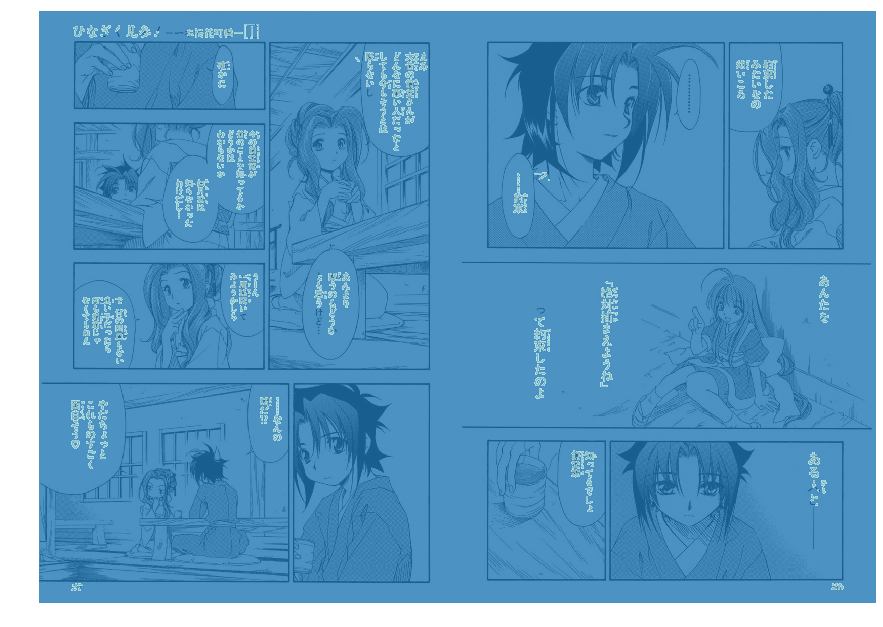}
\caption{Original image with prediction after applying noise removal}
\label{post-processing-3}
\end{figure}

This allows small contours near letters to be kept, as sometimes a letter was not fully predicted and is broken into several contours. It also allows actual dots that tend to be near other letters be kept as well. All the rest, most of which are usually noise, get removed. An example can be seen in Figs. \ref{post-processing-1}, \ref{post-processing-2} and \ref{post-processing-3}. The main problem is when all letters of the dialogue are dots, no good contour is near them and they get discarded.

\begin{minipage}{\linewidth}
\begin{lstlisting}[language=Python,label={noise}, caption={Noise Removal Algorithm}]
im = mask.permute(1,2,0).numpy() * 255
original = im.copy()
simple = cv2.CHAIN_APPROX_SIMPLE
cnts, _ = cv2.findContours(im.astype('uint8'), cv2.RETR_LIST, simple)
goods = [cv2.contourArea(c) >= 100 for c in cnts]
rects = [cv2.boundingRect(c) for c in cnts]
changed = True
ran = range(0, len(cnts)
while changed:
    changed = False
    for c, good, idx, rect in zip(cnts, goods, rang), rects):
        x,y,w,h = rect
        x, y = x + w / 2, y + h / 2 
        if good:
            continue
        for a in range(max(idx - 15, 0), min(idx + 15, len(cnts))):
            if a != idx and goods[a]:
                x2, y2, w2, h2 = rects[a]
                x2, y2, = x2 + w2 / 2, y2 + h2 / 2 
                closeInY = abs(y2 - y) < (h + h2) / 2 + 10
                closeInX = abs(x2 - x) < (w + w2) / 2 + 20

                if closeInY and closeInX:
                    good = goods[idx] = True
                    changed = True
                    break
                        

for c, good, _, _ in zip(cnts, goods, rang), rects):                        
    if not good:
        cv2.drawContours(im, [c], 0, (0, 0, 0), -1)
ImageSegment(tensor(im).permute(2,0,1))
\end{lstlisting}
\end{minipage}

\clearpage

Noise removal helps removing mistakes in the prediction, with a small risk of removing objects that were correct. Another problem to try to solve is to fix partial predictions (letters not completely covered). This means trying to detect cases where a letter was not completely covered by the prediction and then expanding that prediction to the whole letter.

Algorithm can be seen in listing \ref{expand}. Image is first converted to gray-scale and then applied an adaptive threshold method to separate into black and white. Then we get the contours, hoping that each letter will have its own contour. We iterate the contours, get the bounding box and further refine the threshold by applying it to the bounding box crop of the gray-scale image. Then we get the connected components of that region. If they are less than 10, we iterate the components. If the component has an area of more than 3 pixels and the intersection of that area with the prediction is greater than 10\% of the area (we predicted at least 10\% of the component to be text), we consider the whole component to be text.

\begin{minipage}{\linewidth}
\begin{lstlisting}[language=Python,label={expand}, caption={Fix Partial Detection Algorithm}]
mask = mask.astype('uint8')
gray = cv2.cvtColor(img, cv2.COLOR_RGB2GRAY)
adapt = cv2.ADAPTIVE_THRESH_GAUSSIAN_C
bin = cv2.THRESH_BINARY
aprox = cv2.CHAIN_APPROX_SIMPLE
cvThres = cv2.adaptiveThreshold
thresh = cvThres(gray,255,adapt, bin,15,30)
cnts, _ = cv2.findContours(thresh, cv2.RETR_LIST, aprox)
im3 = np.zeros(thresh.shape, np.uint8)

for c in cnts:
    x,y,w,h = cv2.boundingRect(c)
    thresh = cvThres(gray[y:y+h, x:x+w],255,adapt, bin,15,30)
    thresh = cv2.bitwise_not(thresh)
    ret, markers = cv2.connectedComponents(thresh, connectivity=8)
    if ret < 10:
        for label in range(1,ret):
            m = markers == label
            intersec = (m & mask[y:y+h, x:x+w] > 0).sum()
            if m.sum() > 3 and intersec > m.sum() * 0.1:
                im3[y:y+h, x:x+w][m] = 255
\end{lstlisting}
\end{minipage}

\begin{figure}[h!]
\center
\includegraphics[height=0.45\textheight]{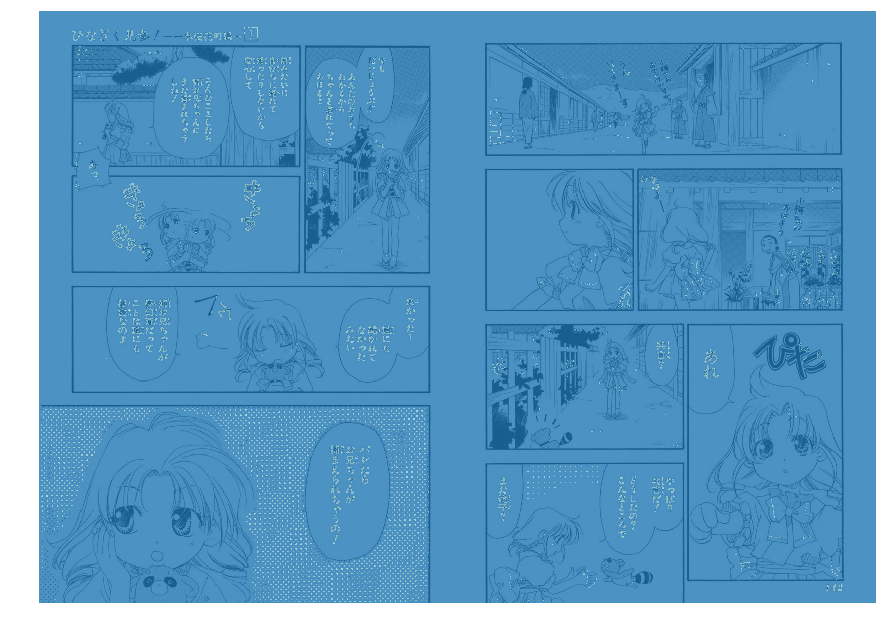}
\caption{Raw Prediction}
\label{expand-1}
\end{figure}

\begin{figure}[h!]
\center
\includegraphics[height=0.45\textheight]{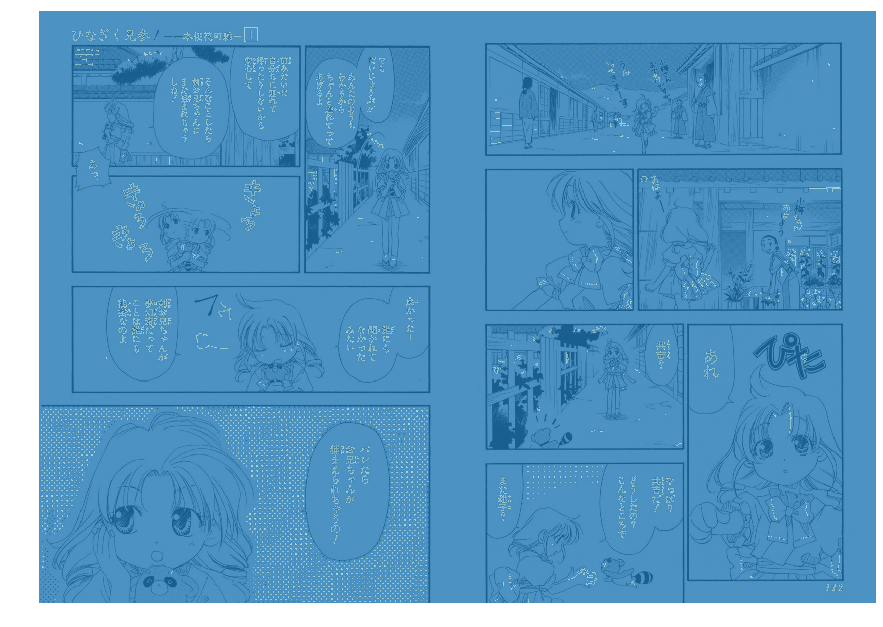}
\caption{Prediction after expanding partial detections}
\label{expand-2}
\end{figure}

\begin{figure}[h!]
\center
\includegraphics[height=0.45\textheight]{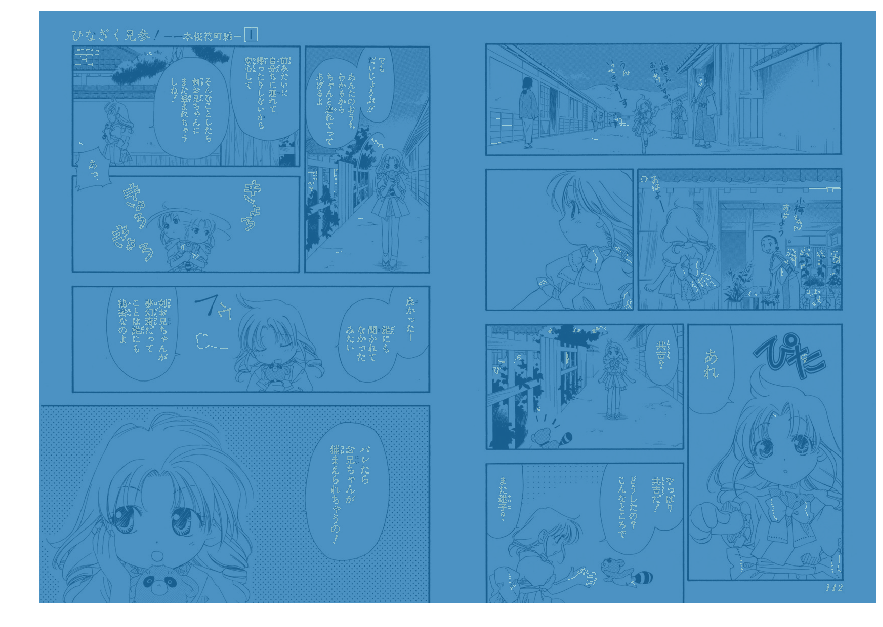}
\caption{Prediction after expanding partial detections and removing noise}
\label{expand-3}
\end{figure}

An example can be seen in Figs. \ref{expand-1}, \ref{expand-2} and \ref{expand-3}.

\section{Conclusion}
After trying out many models, hyper-parameters and modifications on how to generate synthetic data, we finally got good results: most of the letters inside speech bubbles are recognized. Still, sound effects are rarely recognized and there were many false positives, specially when images were full of small dots. The dot issue was mostly fixed with a noise removal algorithm, but the non detection of sound effects is more challenging, as generating those synthetically is hard. Therefore, we conclude our experiments on synthetic data here, having reached good results and solved most of the problems. 

\chapter{Segmentation on real images}
\section{Dataset}\label{dataset}
As previously stated, we had solved most of the problems in detecting text in manga. However, the issue of non standard characters such as sound effects were still an issue. In order to try to solve this, we decided to train on real images. 

There are very few datasets of images with text and their corresponding pixel level mask. This is mainly due to the large amount of time required to label them properly. Some of them are: \texttt{ICDAR} (2013) \cite{icdar2013}, \texttt{Total-Text} (2018) \cite{total_text} and \texttt{COCO\_TS} (2019) \cite{bonechi2019cocots}. However, most of them correspond to real-world images, which differ greatly from manga.

We tried making a datasets of synthetic images using manga-style images (Daanbooru2019) without text and adding text to them of a particular font and size. However, randomly adding text characters anywhere does not replicate where the text is naturally placed in manga, as much text is inside speech bubbles and near characters. 

Synthetically replicating the speech balloons is not easy either, as they are not always a simple rectangle or ellipse like shape. Besides, text outside speech balloons are part of the artwork, and usually have unique artistic styles of the author.

\texttt{Manga109} \cite{manga109}\cite{Matsui_2016}  is the largest public manga dataset, providing bounding boxes for many types of objects, including text. However, it does not have pixel-level masks, and not all text has a bounding box. 

Taking into account all these issues, we decided to create our own dataset with pixel-level annotations. We chose to use images from \texttt{Manga109}, as it is a known public dataset, features a wide range of genres and styles, and the manga authors have granted permission to use and publish their works for academic research. To cover as many different styles as possible, few images from many manga volumes are preferable to a lot from few volumes, as long as those few are enough for the network to learn its style. After observing many examples, we concluded that the first ten images of each manga volume in the \texttt{Manga109} dataset were a suitable number, as that included the cover of the manga and a few pages of the actual content. Thus we manually annotated with pixel-level text masks
the first ten images from 45 different digital mangas, 
 totalizing 450 images. We used photoshop and GIMP for this task. Depending on the amount and style of text, most images took between 20 and 40 minutes each.  
As each manga image in the  \texttt{Manga109} dataset corresponds to 2 pages of a physical manga, we digitally annotated 900 physical pages of mangas. 

\begin{figure}[h]
\centering
\begin{subfigure}[b]{0.45\textheight}
   \includegraphics[width=\linewidth]{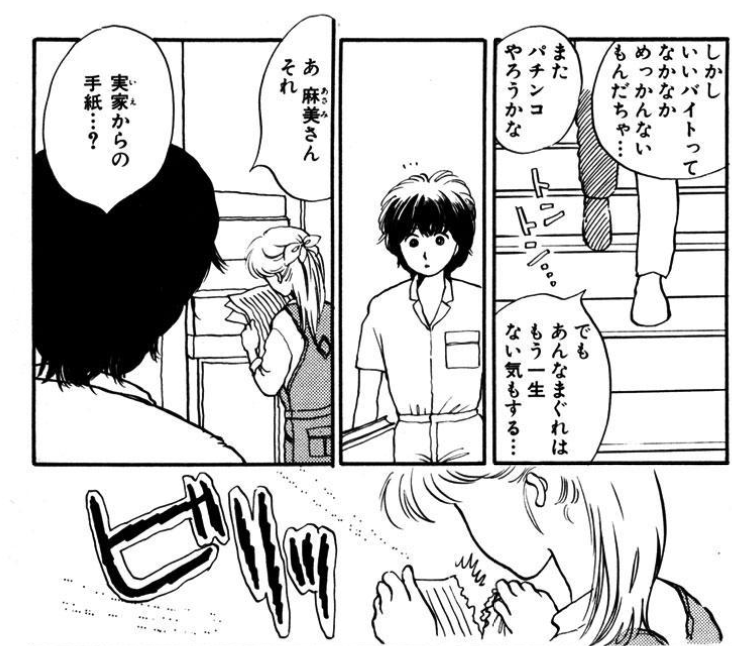}
            \caption{}
\end{subfigure}
\begin{subfigure}[b]{0.45\textheight}
   \includegraphics[width=\linewidth]{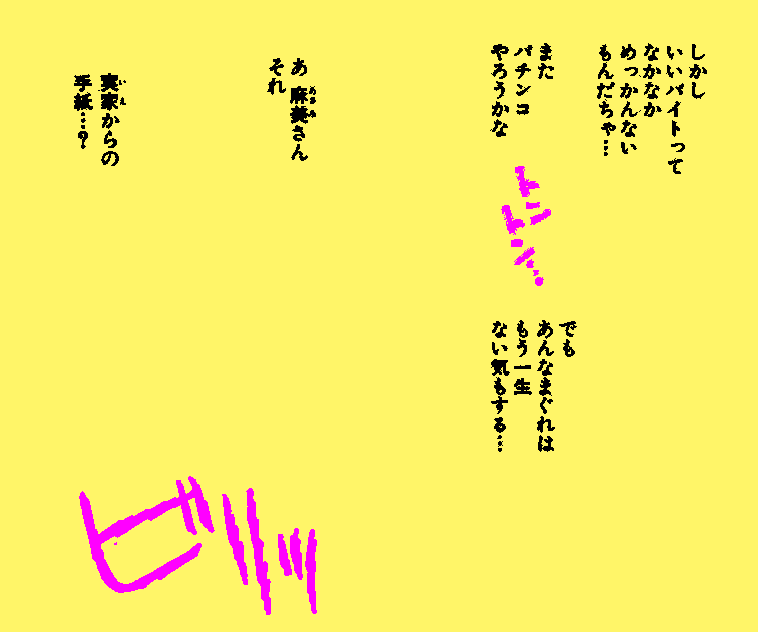}
            \caption{}
\end{subfigure}
   \caption{
   Example of segmentation mask in our dataset. (a) Original image. Speech text is usually inside balloons and sound effects outside. Note the sound effects, near the stairs, and ripped paper. 
   Image from ``Aisazu Niha Irarenai'' \textcopyright {Yoshi Masako},  \texttt{Manga109} dataset \cite{manga109}\cite{Matsui_2016}\cite{ogawa2018object}.
   (b) Corresponding segmentation mask in our dataset. The text inside speech balloons is considered as an easy detection task and labeled with black. Text outside balloons is considered a difficult text detection and labeled with pink. Non-text pixels are labeled with yellow
   }
\label{fig:example1}
\end{figure}

Instead of a simple binary mask (text and non-text), we label the dataset with 3 classes (Fig. \ref{fig:example1}, b): non-text, easy text (text inside speech balloons), and hard text (text outside speech balloons). While we still use the binary version for training, we use this separation of difficulties on text characters for a better understanding of model performance in metric evaluation.

While labeling each character with different colors would be ideal and also useful for text recognition, it is an extremely time consuming task and requires more knowledge about Japanese characters to be able to differentiate them. The last time we are aware that a pixel character labeling was done was with ICDAR 2013, featuring about 500 images. As labeling this kind of data is too expensive, models have been improved over time to not depend on this and work well enough with polygon or bounding boxes around words and their corresponding transcripts.

\section{Evaluation Metrics}\label{metrics}

\begin{figure}[h]
\begin{center}
   \includegraphics[width=0.42\linewidth]{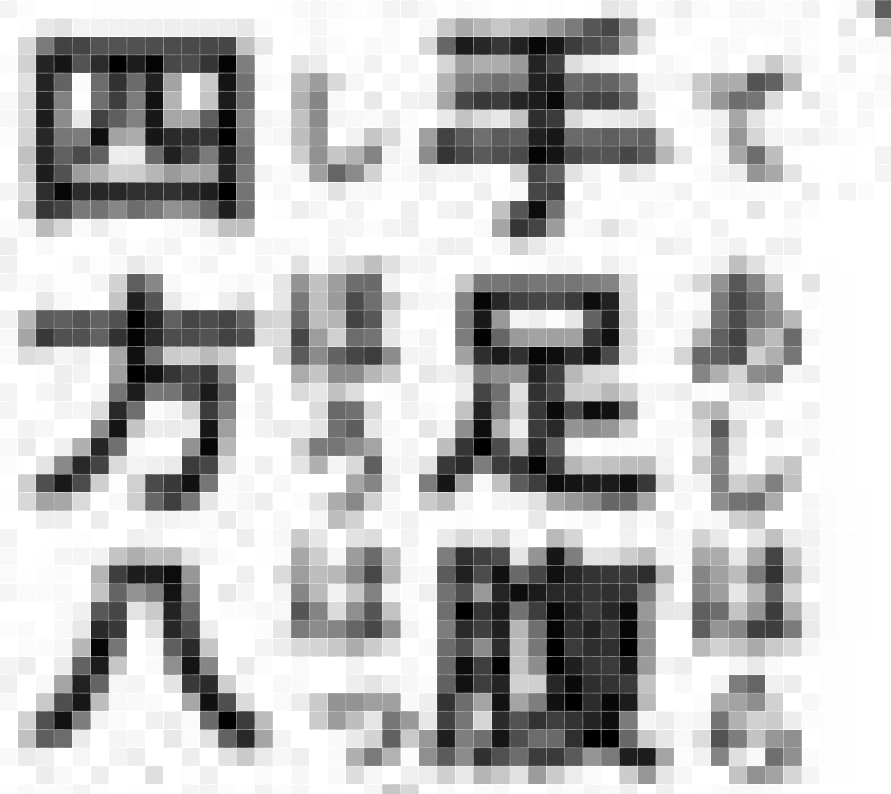}
\end{center}
   \caption{Example of text inside the speech bubble zoomed in. Note that text boundary is unclear and prone to error due to artifacts. 
   Image from ``Akkera Kanjinchou'' \textcopyright {Kobayashi Yuki},  \texttt{Manga109} dataset \cite{manga109}\cite{Matsui_2016}\cite{ogawa2018object}   
   }
\label{fig:blurring}
\end{figure}

Metrics such as recall, precision, F$_1$ score, and dice at a pixel level are commonly used to evaluate binary segmentation models in images. These assume the data is perfectly labeled and allow no compromises on the boundary, which is the part most prone to error. In many tasks, such as segmenting vehicles, this doesn’t matter much as the area of a car is very big compared to the area that might be wrongly labeled, so the human error in labeling won’t account much to influence metrics. With text, however, this is not the case. Not only are characters usually small, but also the boundary is many times unclear because of artifacts and blurring, as noted in Fig. \ref{fig:blurring}. Another issue is that a large text character can have the same area as 100 small characters, making a model that correctly matches most of its pixels but none of the other 100 characters, as good as one matching the 100 small ones but little of the big one.

\begin{figure}[h]
\centering
    \begin{subfigure}[b]{0.3\textwidth}
        \includegraphics[width=\textwidth]{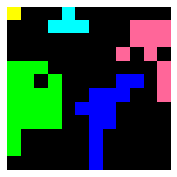}
        \caption{Ground Truth}
        \label{fig:watersheda}
    \end{subfigure}
    \begin{subfigure}[b]{0.3\textwidth}
        \includegraphics[width=\textwidth]{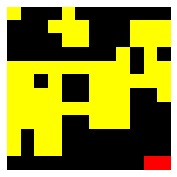}
        \caption{Prediction}
        \label{fp}
    \end{subfigure}
    \begin{subfigure}[b]{0.3\textwidth}
        \includegraphics[width=\textwidth]{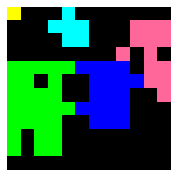}
        \caption{Watershed result}
    \end{subfigure}    
   \caption{ Example of the watershed algorithm matching prediction pixels to ground truth connected components. (a) Masks of five ground truth connected components (labeled in different colors to be distinguished); (b) Predicted text segmentation mask. We marked in red (bottom, right) a predicted connected component with no correspondence to any ground truth connected component; (c) The predicted mask is matched with the ground truth using the watershed algorithm to obtain the evaluation metrics. Five detections matching ground truth connected components are obtained}
\label{fig:watershed}
\end{figure}

Calarasanu \etal \cite{calarasanu.phd}\cite{calarasanu.16.iwrr}\cite{calarasanu} have proposed several metrics to account for these issues. In this work, we have adopted an approach similar to theirs. In addition to the standard pixel metrics, we calculate metrics based on connected components. A connected component in these images is a region of adjacent pixels, considering its 8  neighbors, sharing the same value (see Fig. \ref{fig:watersheda}). 

Given a ground truth connected component $G_{i}$ and its matching detection $D_{j}$, its accuracy and coverage are defined as: 

\begin{align}
 Acc_i=\frac{Area(G_{i}\bigcap D_{j})}{Area(D_{j})} \label{eq:accuracy}\\
 Cov_i=\frac{Area(G_{i}\bigcap D_{j})}{Area(G_{i})} 
\end{align}

To account for multiple detections matching a single ground truth or a single detection matching multiple ground truths, we apply the watershed algorithm to match prediction pixels to a single ground truth, as seen in Fig. \ref{fig:watershed}. 

We define $tp$ as the number of ground truth connected components that have at least one pixel of detection associated with it. We define $fp$ as the number of detected connected components which had no correspondence to any ground truth (see Fig. \ref{fp}).

Given a dataset with $m$ ground truth connected components and $d$ detections, 
quantity recall $R_{quant}$ and quantity precision $P_{quant}$ are defined as:

\begin{align}
 R_{quant}=\frac{tp}{m} \\
 P_{quant}=\frac{tp}{tp + fp}
\end{align}

Quality recall $R_{qual}$, quality precision $P_{qual}$ and $F1_{qual}$ are defined as:

\begin{align}
 R_{qual}=\frac{\sum_{i=0}^{tp} Cov_i}{tp} \\
 P_{qual}=\frac{\sum_{i=0}^{tp} Acc_i}{tp} \\
 F1_{qual}= \frac{2\  R_{qual}\ P_{qual}}{ R_{qual} + P_{qual}} \label{f1qual}
\end{align}

Global recall $GR$, global precision $GP$,  and global F$_1$ $GF1$ are defined as:
\begin{align}
 GR=R_{quant} R_{qual}=\frac{\sum_{i=0}^{tp} Cov_i}{m} \\
 GP=P_{quant} P_{qual}=\frac{\sum_{i=0}^{tp} Acc_i}{tp + fp} \\
 GF1= \frac{2\ GR\ GP}{GR + GP}
\end{align}

Standard metrics for pixels are defined as:
\begin{align}
Precision=\frac{TP}{TP+FP}
\end{align}

\begin{align}
Recall=\frac{TP}{TP+FN}
\end{align}

\begin{align}
 Pixel F1= PF1 = \frac{2\ Recall\ Precision}{Recall + Precision}
\end{align}

With TP being pixels that were correctly segmented as text (true positive), FP being pixels that were wrongly segmented as text (false positives) and FN being pixels that were wrongly segmented as background (false negatives).

\begin{figure}[h]
\centering
    \begin{subfigure}[b]{0.3\textwidth}
        \includegraphics[width=\textwidth]{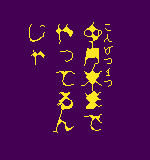}
        \caption{Eroded GT}
    \end{subfigure}
    \begin{subfigure}[b]{0.3\textwidth}
        \includegraphics[width=\textwidth]{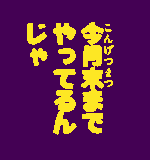}
        \caption{Original GT}
    \end{subfigure}
    \begin{subfigure}[b]{0.3\textwidth}
        \includegraphics[width=\textwidth]{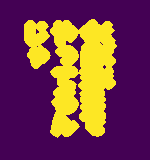}
        \caption{Dilated GT}
    \end{subfigure}    
   \caption{Example of segmentation masks used in normal mode and  relaxed mode metrics. (a) The eroded mask under-segments the ground truth mask. It is used for coverage in relaxed mode; (b) Ground truth mask. It is used in normal mode; (c) The dilated mask over-segments the ground truth. It is used for accuracy in relaxed mode. In relaxed mode, if the network predicts the ground truth of b), it has 100\% of accuracy. However, if the prediction is inside the dilated mask of c), it also has 100\% of accuracy. Accuracy is measured with Equation \ref{eq:accuracy}.  }
\label{fig:dilation_erosion}
\end{figure}

We calculate metrics in  normal and relaxed mode. Normal mode assumes that the dataset is perfectly labeled. Relaxed mode tries to lessen the effect of wrong boundary labeling (Fig. \ref{fig:dilation_erosion}). In normal mode, we calculate the metrics using the segmentation masks of the dataset without modification.  In relaxed mode, an eroded version of the ground truth is used to calculate coverage while a dilated version is used to calculate accuracy. In both modes, we consider there is no match to a ground truth component when there is no intersection between the eroded version and prediction, as the eroded version is the most important part to detect. For both erosion and dilation, a cross-shaped structuring element is used (connectivity=1).

\clearpage

\section{Methodology}\label{model}
Our text detector model employs a U-net  \cite{unet} architecture with a pre-trained resnet34 \cite{He2015DeepRL} backbone. Despite having been pre-trained with ImageNet, which features images quite different from  manga, it has proved to work well. We implemented the model in PyTorch. We used the \texttt{fastai} U-Net model \cite{fastaiunet}. We  trained the network with the \texttt{fastai} library \cite{fastai}\cite{fastaip}, making use of its one cycle policy, a modified version of the one initially devised by Leslie N. Smith \cite{smith}. The encoder part was frozen, and only the decoder part was trained, as the encoder already comes with the pre-trained weights from ImageNet. As we handle binary segmentation, a single channel is used as the last layer to provide the logits of a pixel being text. We later apply a sigmoid function and set 0.5 as a threshold to consider whether to classify it as text or background. As for the loss function, dice loss is used, which showed considerably better results than the simple binary cross-entropy loss (see Section \ref{loss}). In the next section, we show how we used our metrics of Section \ref{metrics} to select an optimal loss function  and an optimal architecture for the model.

The images of the  \texttt{Manga109} dataset are 1654 width and 1170 height. As they represent sheets of paper from physical books, in almost all cases (with some covers as the exception), the two pages from it have no text in the middle and can be split without affecting text characters. We took advantage of that and cut the images of our dataset in half, so we end up with 900 manga pages to train (see Section \ref{dataset}). The only data augmentation used is a 512x800 random crop for training. We tried a few other data augmentations such as flip and warp, but we didn’t notice any significant improvement.

We used K-Fold cross-validation with five folds to calculate all metrics, leaving 20\% as validation. We show the validation folders used for each split in Table \ref{table:folds}. Between transfer learning, one cycle policy, and a batch size of 4, results are obtained by training for ten epochs, which is completed in less than an hour on a single GeForce GTX 1080 Ti GPU for a single fold.
\begin{table}[h!]
\centering
\begin{tabular}{| c | c | c |}
\hline
Split 1 & Split 2 & Split 3\\
\hline
AosugiruHaru & UnbalanceTokyo & UchiNoNyan'sDiary \\
TouyouKidan & Akuhamu & Belmondo \\
HanzaiKousyouninMinegishiEitarou & Arisa & HarukaRefrain \\
HaruichibanNoFukukoro & ToutaMairimasu & AppareKappore \\
BakuretsuKungFuGirl & Hamlet & UltraEleven \\
YasasiiAkuma & Count3DeKimeteAgeru & DualJustice \\
ByebyeC-BOY & Donburakokko & UchuKigekiM774 \\
YumeNoKayoiji & EienNoWith & YoumaKourin \\
TotteokiNoABC & EverydayOsakanaChan & ARMS \\
\hline
\noalign{\vskip 5mm}    
\end{tabular}
\begin{tabular}{ | c | c |}
\hline
Split 4 & Split 5 \\
\hline
YamatoNoHane & TsubasaNoKioku \\
BurariTessenTorimonocho & HealingPlanet \\
GakuenNoise & DollGun \\
YumeiroCooking & GarakutayaManta \\
AisazuNihaIrarenai & BEMADER\_P \\
AkkeraKanjinchou & BokuHaSitatakaKun \\
GinNoChimera & YukiNoFuruMachi \\
YouchienBoueigumi & EvaLady \\
WarewareHaOniDearu & GOOD\_KISS\_Ver2 \\
\hline
\end{tabular}
\caption{Different folds used in experiments. Each fold represents the folders used for validation, with each folder having the first 10 images of the \texttt{Manga109} dataset}
\label{table:folds}
\end{table}

\section{Experiments}\label{experiments}

\FloatBarrier
\subsection{Loss Function Selection}\label{loss}
Choosing an adequate loss function is a crucial step in the design of a machine learning model. In this section, we show how we used our metrics of Section \ref{metrics} to find a suitable loss function.

We used our dataset to train a U-net network, which is a model commonly used in segmentation, with different loss functions. Then, we used the metrics of section \ref{metrics} to measure the performance of the model for each different loss function.
For the experiments we used the following loss functions:
binary cross-entropy (BCE), which is a loss function commonly used in binary segmentation, and a mixed loss  that combines focal loss \cite{FocalLF} and dice loss, and is defined as,

\begin{align}
Mix(\alpha, \gamma)=\alpha FocalLoss(\gamma) - log(DiceLoss) \label{formu}
\end{align}
We trained a \texttt{fastai resnet18} U-net with each loss function  during ten epochs, with $0.001$ as the maximum learning rate. 
Table \ref{table:loss} shows the metrics averaged over 5 folds varying the parameters of the loss functions.
\begin{table}[h!]
\centering
 \caption{Metrics averaged over 5 folds for a \texttt{fastai resnet18}  U-Net with different loss functions. We show the mean and standard deviation of the metrics. PF1 indicates standard Pixel F1 score, while GF1, P$_{quant}$ takes into account connected components as defined in section \ref{metrics} 
 }
\begin{tabular}{ |c|c|c|c|c|c|}
 \hline
 \multirow{2}{*}{Loss} & \multicolumn{2}{c|}{Normal} & \multicolumn{2}{c|}{Relaxed} & Both \\
 \cline{2-6}
 & PF1 & GF1 & PF1 & GF1 & $P_{quant}$\\
 \hline
BCE(0.5) & $67.99 \pm 1.2$ & $63.79 \pm 0.9$ & $71.30 \pm 2.3$ & $73.32 \pm 0.9$ & $71.01 \pm 2.4$  \\
BCE(1) & $70.97 \pm 1.1$ & $63.09 \pm 1.2$ & $75.31 \pm 1.9$ & $72.13 \pm 1.5$ & $65.06 \pm 3.0$  \\
BCE(5) & $65.26 \pm 2.5$ & $53.39 \pm 2.8$ & $73.38 \pm 2.7$ & $64.54 \pm 1.8$ & $51.79 \pm 2.6$  \\
BCE(10) & $59.03 \pm 3.1$ & $46.34 \pm 2.9$ & $68.91 \pm 3.5$ & $59.11 \pm 1.8$ & $45.48 \pm 1.8$  \\
BCE(30)  & $48.64 \pm 2.5$ & $35.44 \pm 3.1$ & $60.65 \pm 3.0$ & $50.28 \pm 2.8$ & $36.86 \pm 2.4$  \\
Mix(5, 1) & \bm{$72.63 \pm 1.7$} & $68.40 \pm 1.4$ & $77.81 \pm 2.3$ & $78.67 \pm 1.0$ & $73.20 \pm 3.3$  \\
Mix(10, 1)  & $71.97 \pm 1.5$ & $63.80 \pm 1.4$ & $77.05 \pm 2.0$ & $73.50 \pm 1.3$ & $65.26 \pm 3.1$ \\
Mix(10, 2) & $71.74 \pm 1.5$ & $62.66 \pm 1.3$ & $76.95 \pm 2.1$ & $72.27 \pm 1.2$ & $63.36 \pm 2.9$ \\
Mix(5, 2) & $72.62 \pm 1.8$ & $68.33 \pm 1.1$ & $77.86 \pm 2.4$ & $78.63 \pm 0.9$ & $73.07 \pm 3.0$  \\
Mix(0, 1)  & \bm{$72.63 \pm 1.8$} & \bm{$71.95 \pm 1.5$} & \bm{$77.89 \pm 2.9$} & \bm{$82.93 \pm 1.4$} & \bm{$79.81 \pm 2.8$}  \\
\hline
\end{tabular}
\label{table:loss}
\end{table}

As seen in Table \ref{table:loss}, the $Mix$ loss function outperformed the BCE in all metrics.
While there is little difference in $PF1$ scores between the different $Mix$ losses, there is a big difference in $GF1$, making $Mix(0,1)=-log(DiceLoss)$ the one with the highest scores in them.  

As $Mix(0, 1)$ showed the best results,
 we chose it as our loss function for further experiments.
 
\FloatBarrier
\subsection{Model Architecture Selection}\label{architecture}

To select an optimal architecture, 
we trained different models with our dataset. The models were trained under the same conditions, 
always using $Mix(0, 1)$ as the loss function. We used the metrics of Section \ref{metrics} to measure the performance of the model and choose an optimal architecture.

We show in Table \ref{table:model} a comparison of \texttt{fastai} U-Net learner \cite{fastaiunet} using $resnet18$ and $resnet34$ \cite{He2015DeepRL} encoders and yu45020's \cite{yu45020} models using $xception$ \cite{Chollet2016XceptionDL} and \textit{mobileNetV2} \cite{mobilenetv2}. For $xception$ and $mobileNetV2$ we trained them from scratch with our labeled dataset. For \texttt{fastai} U-Net $resnet18$ and $resnet34$, only the decoder was trained.

\begin{table}[h!]
\centering
 \caption{Metrics after training different architectures}
\begin{tabular}{ |c|c|c|c|c|  }
 \hline
 \multirow{2}{*}{Model} & \multicolumn{2}{c|}{Normal} & \multicolumn{2}{c|}{Relaxed} \\
 \cline{2-5}
 & PF1 & GF1 & PF1 & GF1 \\
 \hline
    \texttt{fastai} resnet18 U-Net & $72.63 \pm 1.8$ & $71.95 \pm 1.5$ & $77.89 \pm 2.9$ & $82.93 \pm 1.4$ \\
    \texttt{fastai} resnet34 U-Net & \bm{$73.91 \pm 2.1$} & \bm{$74.65 \pm 1.2$} & \bm{$78.65 \pm 3.4$} & \bm{$85.70 \pm 1.3$} \\
    mobileNetV2 & $47.97 \pm 5.4$ & $39.17 \pm 7.2$ & $50.54 \pm 6.3$ & $52.26 \pm 8.4$ \\
    xception & $61.61 \pm 5.6$ & $62.48 \pm 3.2$ & $67.48 \pm 7.1$ & $79.35 \pm 4.2$ \\
  \hline
\end{tabular}
 \label{table:model}
\end{table}

We also experimented with several segmentation models from \texttt{qubvel}'s library \cite{qubvel}, training only the decoder and using default parameters. However, the \texttt{fastai} U-Net learner outperformed the models in this library in all metrics for this problem. As the training was more unstable, we calculated top metrics during training instead of the final score after the last epoch, taking epoch with the highest relaxed $GF1$ score. Results can be seen in Table \ref{table:qubvel}. It is interesting to note that the U-net architecture worked better in terms of relaxed $GF1$ score in all qubvel’s encoders.

\begin{table}[h!]
\centering
 \caption{Top scores during training of diverse architectures from \texttt{qubvel}'s library}
\begin{tabular}{ |c|c|c|c|c|  }
 \hline
\multirow{2}{*}{Loss} & \multicolumn{2}{c|}{Normal} & \multicolumn{2}{c|}{Relaxed} \\
 \cline{2-5}
 & PF1 & GF1 & PF1 & GF1 \\
 \hline
densenet169 FPN  & $55.40 \pm 3.0$ & $51.04 \pm 1.1$ & $70.87 \pm 1.1$ & $71.09 \pm 2.2$ \\
densenet169 Linknet  & \bm{$60.74 \pm 1.3$} & $53.85 \pm 1.3$ & \bm{$77.38 \pm 2.1$} & $76.71 \pm 1.3$ \\
densenet169 PSPNet & $37.34 \pm 12.0$ & $34.38 \pm 4.7$ & $44.12 \pm 12.0$ & $43.28 \pm 6.3$ \\
densenet169 U-Net & $60.60 \pm 2.1$ & $56.25 \pm 1.4$ & $75.93 \pm 1.9$ & $77.36 \pm 1.9$ \\
dpn68 FPN & $53.37 \pm 2.5$ & $50.80 \pm 1.0$ & $68.99 \pm 2.1$ & $70.08 \pm 4.8$ \\
dpn68 Linknet & $42.53 \pm 1.4$ & $36.12 \pm 3.3$ & $61.63 \pm 1.6$ & $61.66 \pm 4.5$ \\
dpn68 PAN  & $56.80 \pm 1.1$ & $48.80 \pm 1.6$ & $69.79 \pm 3.5$ & $65.10 \pm 3.1$ \\
dpn68 PSPNet & $8.27 \pm 1.2$ & $30.78 \pm 1.4$ & $12.01 \pm 1.6$ & $45.59 \pm 1.8$ \\
dpn68 U-Net & $60.42 \pm 1.5$ & $55.19 \pm 2.1$ & $72.97 \pm 2.8$ & $73.42 \pm 3.6$ \\
efficientnet-b4 & $58.60 \pm 1.4$ & $48.08 \pm 0.3$ & $73.15 \pm 3.3$ & $66.12 \pm 0.7$ \\
efficientnet-b4 Linknet & $10.36 \pm 2.6$ & $4.63 \pm 1.3$ & $15.94 \pm 3.7$ & $9.49 \pm 2.2$ \\
efficientnet-b4 PAN & $52.65 \pm 1.3$ & $47.74 \pm 1.1$ & $66.24 \pm 3.3$ & $66.06 \pm 3.3$ \\
efficientnet-b4 PSPNet  & $36.88 \pm 2.7$ & $43.59 \pm 1.4$ & $41.16 \pm 3.3$ & $61.16 \pm 2.0$ \\
efficientnet-b4 U-Net & $60.37 \pm 2.6$ & $55.61 \pm 1.3$ & $75.63 \pm 1.4$ & $76.54 \pm 1.0$ \\
resnet50 FPN & $56.59 \pm 2.1$ & $51.51 \pm 0.6$ & $71.15 \pm 1.7$ & $70.48 \pm 1.3$ \\
resnet50 Linknet & $56.61 \pm 1.5$ & $53.04 \pm 1.9$ & $74.34 \pm 1.5$ & $77.05 \pm 0.9$ \\
resnet50 PAN & $55.04 \pm 3.1$ & $48.67 \pm 2.5$ & $69.38 \pm 2.7$ & $67.95 \pm 1.9$ \\
resnet50 PSPNet & $34.00 \pm 11.5$ & $33.11 \pm 2.8$ & $40.23 \pm 11.6$ & $42.24 \pm 2.4$ \\
resnet50 U-Net & $56.96 \pm 1.7$ & $55.04 \pm 1.8$ & $73.30 \pm 1.8$ & $78.83 \pm 1.6$ \\
vgg16 FPN & $56.36 \pm 1.5$ & $48.92 \pm 0.6$ & $68.63 \pm 4.5$ & $67.57 \pm 0.7$ \\
vgg16 Linknet & $56.70 \pm 1.2$ & $42.28 \pm 2.5$ & $72.04 \pm 2.6$ & $63.33 \pm 3.0$ \\
vgg16 PSPNet & $46.67 \pm 3.7$ & $45.58 \pm 1.2$ & $55.84 \pm 7.9$ & $58.54 \pm 1.9$ \\
vgg16 U-Net & $60.48 \pm 0.9$ & \bm{$57.70 \pm 1.0$} & $75.17 \pm 2.7$ & \bm{$80.12 \pm 0.9$} \\
  \hline
\end{tabular}
 \label{table:qubvel}
\end{table}

We chose the best two architectures from previous experiments (\texttt{fastai} resnet18  U-net and \texttt{fastai} resnet34   U-net) and trained  them for five more epochs with all layers unfrozen. As seen in table \ref{table:refinement}, there was a slight improvement in metrics. 

\begin{table}[h!]
\centering
 \caption{First refinement with all layers unfrozen}
\begin{tabular}{ |c|c|c|c|c|  }
 \hline
 \multirow{2}{*}{Model} & \multicolumn{2}{c|}{Normal} & \multicolumn{2}{c|}{Relaxed} \\
 \cline{2-5}
 & PF1 & GF1 & PF1 & GF1 \\
 \hline
    \texttt{fastai} resnet34 U-Net & \bm{$76.44 \pm 1.8$} & \bm{$76.97 \pm 0.6$} & \bm{$81.08 \pm 2.7$} & \bm{$88.00 \pm 1.1$} \\
    \texttt{fastai} resnet18 U-Net & $74.71 \pm 2.2$ & $75.40 \pm 0.5$ & $79.05 \pm 3.8$ & $86.30 \pm 1.0$ \\
  \hline
\end{tabular}
 \label{table:refinement}
\end{table}

 We further refined the network by training for three more epochs without random crop and using the whole image instead of half, reducing the batch size to 1. As seen in Table \ref{table:refinement2}, there was another slight improvement in metrics. 

\begin{table}[h!]
\centering
 \caption{Final refinement with full size image input}
\begin{tabular}{ |c|c|c|c|c|  }
 \hline
 \multirow{2}{*}{Model} & \multicolumn{2}{c|}{Normal} & \multicolumn{2}{c|}{Relaxed} \\
 \cline{2-5}
 & PF1 & GF1 & PF1 & GF1 \\
 \hline
    \texttt{fastai} resnet34 U-Net & \bm{$79.36 \pm 1.1$} & \bm{$84.92 \pm 1.0$} & \bm{$80.43 \pm 1.8$} & \bm{$89.26 \pm 1.0$} \\
    \texttt{fastai} resnet18 U-Net  & $75.82 \pm 2.4$ & $83.29 \pm 1.6$ & $76.87 \pm 2.7$ & $87.50 \pm 1.2$ \\
  \hline
\end{tabular}
 \label{table:refinement2}
\end{table}

\FloatBarrier
\subsection{Comparison against similar works}
We kept the \texttt{fastai} resnet34 U-net as our model, which has the top metric scores (see Section \ref{architecture}). In this section, we compare the performance of our model against similar previous works (see Section \ref{relatedwork}). We first compare our model against two recent similar works found on Github that make pixel-level text segmentation in manga. One is called \enquote{Text Segmentation and Image Inpainting} by yu45020 \cite{yu45020} and the other \enquote{SickZil-Machine} by KUR-creative \cite{Sickzil}.

Next, as the aim of our method is to detect unconstrained text in Japanese manga, we also compared our work against one of the recent many state-of-the-art models in unconstrained text detection: Character Region Awareness for Text Detection (CRAFT) \cite{baek2019character}. 

Performance metrics against  \texttt{yu45020}'s $xception$ \cite{yu45020} and \textit{SickZil-Machine} \cite{Sickzil} are shown in Table \ref{table:similar_works}. 

Our method shows a definite improvement over all F$_1$ metrics, especially on normal mode.  
For $GF1$ in relaxed mode, the difference is smaller as the penalty for over-segmentation and under-segmentation decreases.  
The big difference between $PF1$ and $GF1$ of \textit{SickZil-Machine} in relaxed mode is caused by many false positives. \textit{SickZil-Machine} tends to do over-segmentation. Thus, the false positive areas tend to be bigger, decreasing its precision in pixel mode but remaining a single connected component regardless of area.

\begin{table}[h]
\centering
 \caption{Performance metrics obtained on our dataset by similar methods under similar conditions. Our method and \texttt{yu45020}'s xception were trained with our \texttt{Manga109} labeled dataset. \textit{SickZil-Machine} author has not released the source code,  but the author has stated that his method was trained with its own  \texttt{Manga109} dataset in which text was labeled at a pixel level. 
 The results shown for \textit{SickZil-Machine} are for an executable program provided by the author. 
 }
\begin{tabular}{ |c|c|c|c|c|  }
 \hline
 \multirow{2}{*}{Author} & \multicolumn{2}{c|}{Normal} & \multicolumn{2}{c|}{Relaxed} \\
 \cline{2-5}
 & PF1 & GF1 & PF1 & GF1 \\
 \hline
\texttt{yu45020}'s xception  & $61.61 \pm 5.6$ & $62.48 \pm 3.2$ & $67.48 \pm 7.1$ & $79.35 \pm 4.2$ \\
SickZil-Machine  & $52.07 \pm 1.9$ & $49.33 \pm 2.1$ & $64.66 \pm 3.5$ & $84.94 \pm 2.0$ \\
Ours & \bm{$79.36 \pm 1.1$} & \bm{$84.92 \pm 1.0$} & \bm{$80.43 \pm 1.8$} & \bm{$89.26 \pm 1.0$} \\
 \hline
\end{tabular}
 \label{table:similar_works}
\end{table}

\begin{figure}[h!]
\centering
 \includegraphics[width=\textwidth]{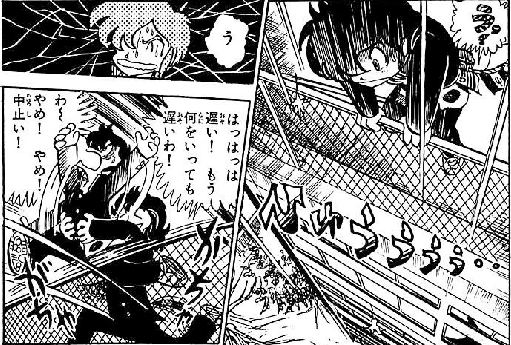}
 \caption{Image  extracted from ``BEMADER\_P'' \textcopyright {Hasegawa Yuichi},  \texttt{Manga109} dataset \cite{manga109}\cite{Matsui_2016}\cite{ogawa2018object}.}
 \label{fig:segmentation-example-1}
\end{figure}

\begin{figure}[h!]
\centering
\begin{subfigure}[t]{0.49\textwidth}
    \includegraphics[width=\textwidth]{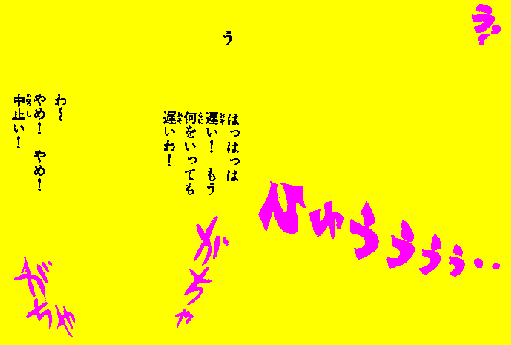}
    \caption{ground truth}
\end{subfigure}
\begin{subfigure}[t]{0.49\textwidth}
    \includegraphics[width=\textwidth]{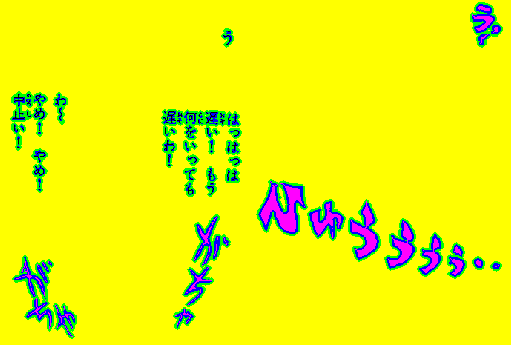}
   \caption{relaxed ground truth}
\end{subfigure}
\begin{subfigure}[t]{0.49\textwidth}
    \includegraphics[width=\textwidth]{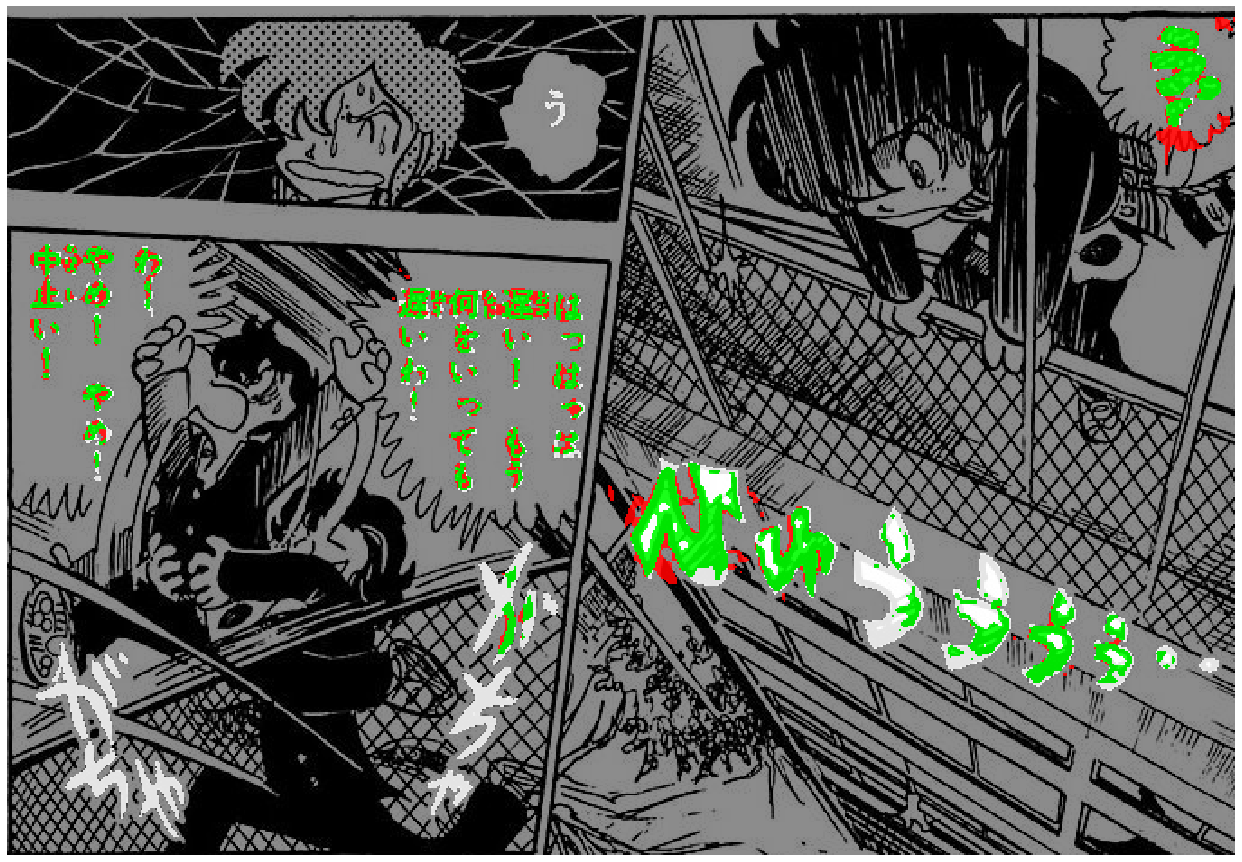}
    \caption{yu45020's xception}
    \label{fig:yumask1}
\end{subfigure}
\begin{subfigure}[t]{0.49\textwidth}
    \includegraphics[width=\textwidth]{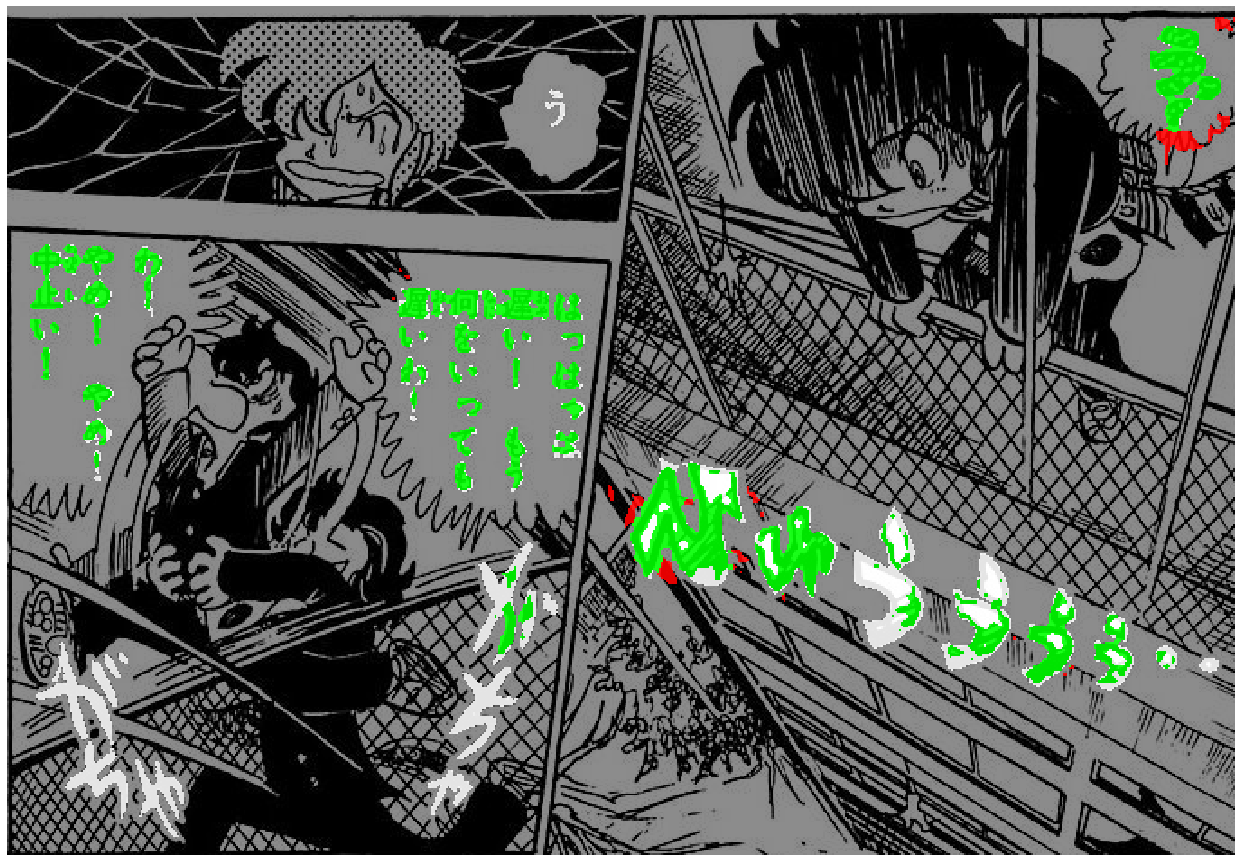}
    \caption{yu45020's xception relaxed}
    \label{fig:yumask2}
\end{subfigure}
\begin{subfigure}[t]{0.49\textwidth}
    \includegraphics[width=\textwidth]{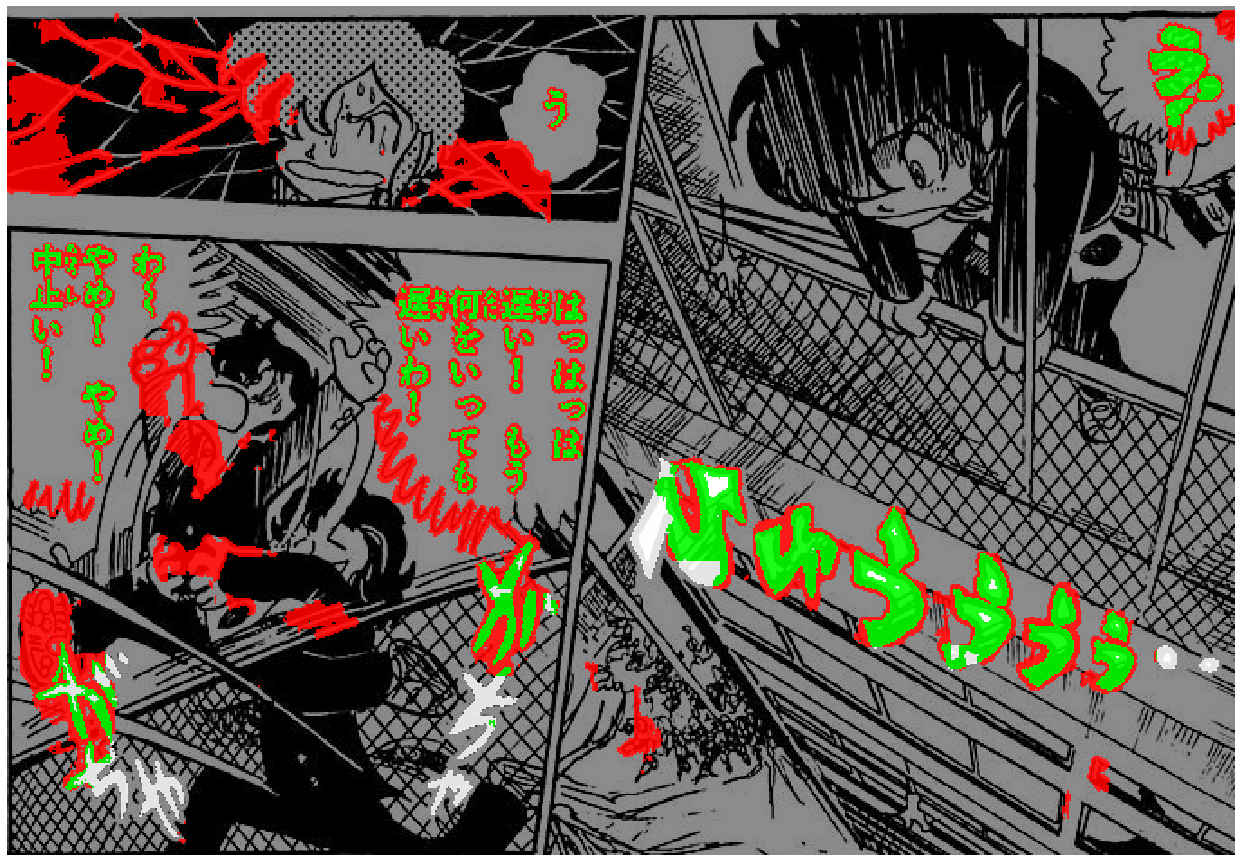}
   \caption{SickZil-Machine}
\label{fig:skmask1}
\end{subfigure}
\begin{subfigure}[t]{0.49\textwidth}
    \includegraphics[width=\textwidth]{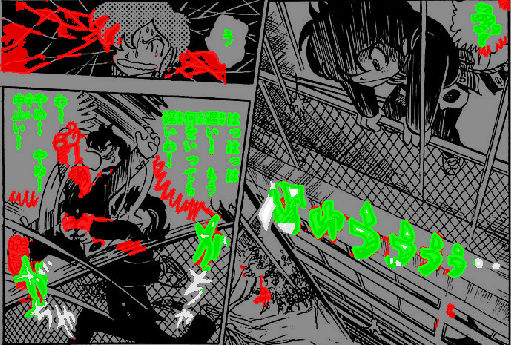}
   \caption{SickZil-Machine relaxed}
\label{fig:skmask2}
\end{subfigure}
\begin{subfigure}[t]{0.49\textwidth}
    \includegraphics[width=\textwidth]{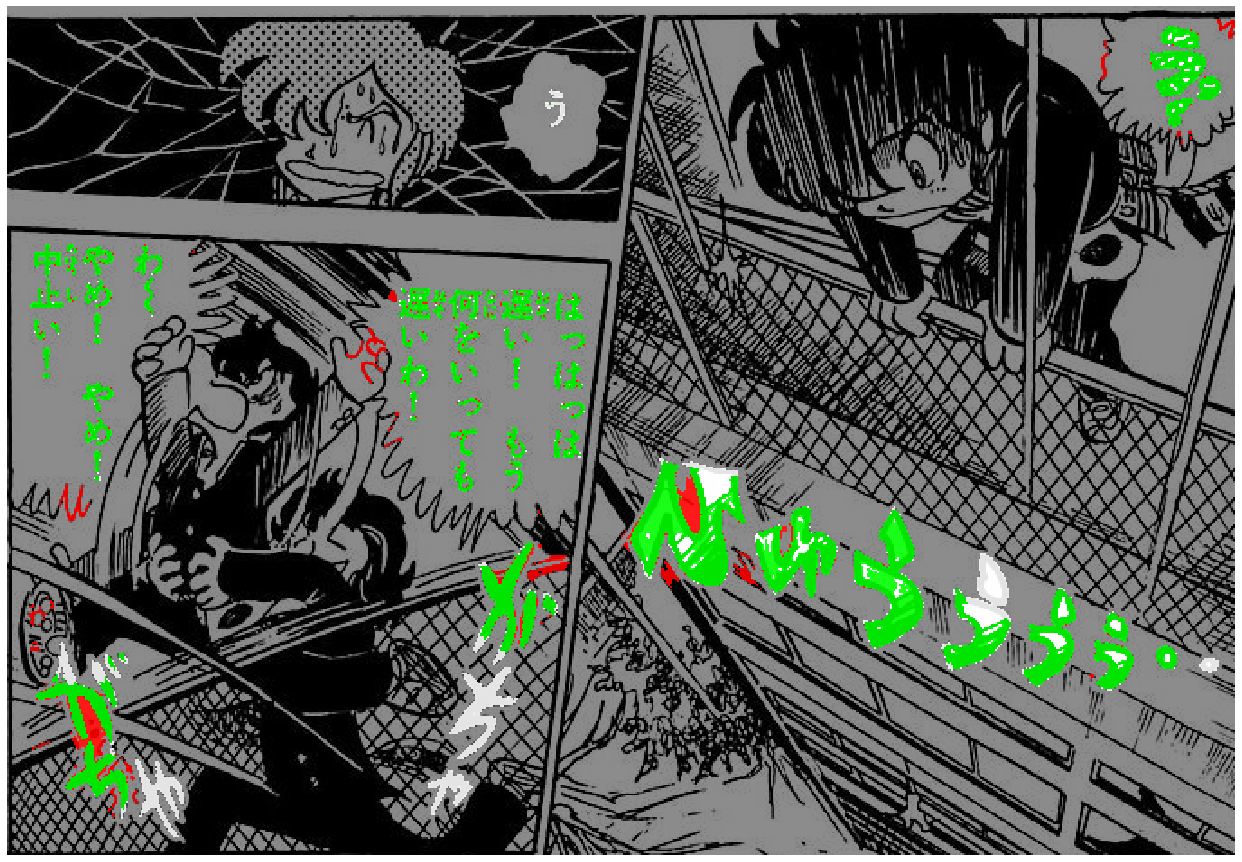}
   \caption{Ours}
    \label{fig:ourmask1}
\end{subfigure}
\begin{subfigure}[t]{0.49\textwidth}
    \includegraphics[width=\textwidth]{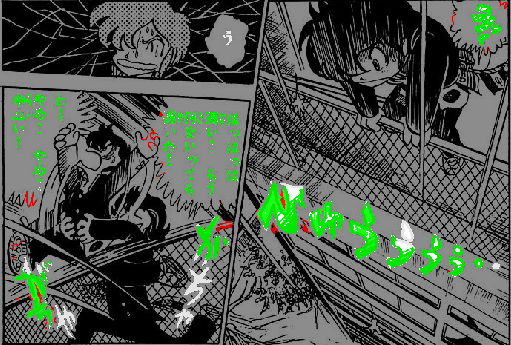}
   \caption{Ours relaxed}
   \label{fig:ourmask2}
\end{subfigure}
\caption{
In red, false positives. In white, missing text. In green, text correctly segmented. (a) Ground Truth. In pink, hard text. In black, easy text. (b) Relaxed Ground Truth. In green, dilated area. In blue, ground truth pixels not belonging to the eroded mask. In (f) boundaries between small letters are lost, but they are still marked as true positives as pixels are inside relaxed dilated area}
\label{fig:similar_works_ours}
\end{figure}

 Fig. \ref{fig:similar_works_ours} shows an example of segmentation masks produced by the different methods over the manga extract seen in Fig \ref{fig:segmentation-example-1}. Our segmentation method misses some of the hard texts but has very few false positives (Figs. \ref{fig:ourmask1}, \ref{fig:ourmask2}). \textit{SickZil-Machine} covers some of those missing texts but also has much more false positives (Figs. \ref{fig:skmask1}, \ref{fig:skmask2}). 
 \texttt{yu45020}’s xception misses many of the hard texts, and detects the small letters with less precision than our model (Figs. \ref{fig:yumask1}, \ref{fig:yumask2}).
 

\begin{figure}[h]
\centering
    \includegraphics[width=\textwidth, height=0.6\textheight]{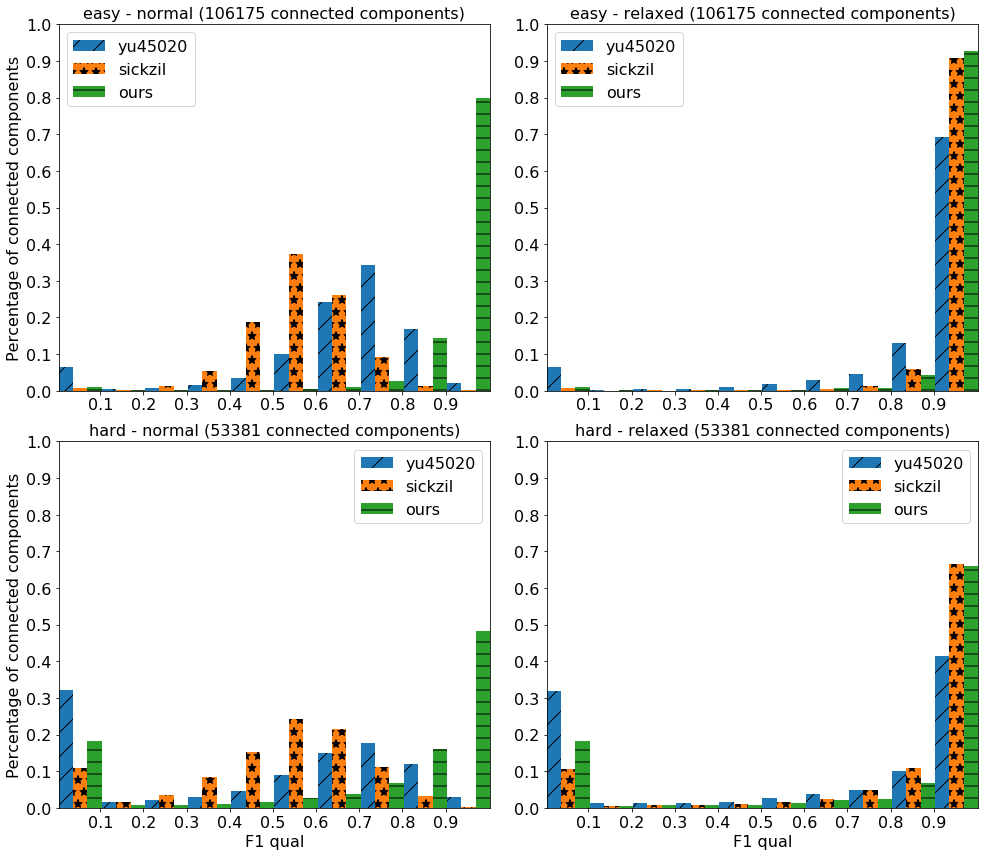}
    \caption{Histogram of F1$_{qual}$ (see Eq. \ref{f1qual}) of the different types of connected components. 
    The first row corresponds with easy text, and the second row corresponds with hard text. The first column corresponds with normal mode, and the second column corresponds with relaxed mode.
    For easy text, our method predicts most of the connected components with a high $F1_{qual}$ value. In normal mode our method has a much higher percentage of easy and hard connected components predicted with a high $F1_{qual}$ value than the other methods.
    }
\label{fig:similar_works_ours_histogram}
\end{figure}

For a global view of the performance on the different types of connected components and segmentation modes, we draw in Fig. \ref{fig:similar_works_ours_histogram} the histograms of $F1_{qual}$ (see Equation \ref{f1qual}). As our method fits the text characters without over-segmentation, it has less false positives, and our method clearly outperforms the other methods for $F1_{qual}$ in normal mode.
For the easy text case in relaxed mode, our method and \textit{SickZil-Machine} detect almost all the connected components.  Thus, we can see that there is little point in adding more data of easy text, as almost all easy components are detected with high F$_1$ score. 

As our method aims to detect unconstrained text, we also compared our work against CRAFT, a scene text detector for unconstrained text \cite{baek2019character}. We used the official implementation, which comes with a trained model for general purposes (General). As the training code is not available for intellectual property reasons, we did not fine tune it with our own manga dataset. 

\begin{figure}[h]
\centering
\begin{subfigure}[t]{\textwidth}
    \includegraphics[width=\textwidth]{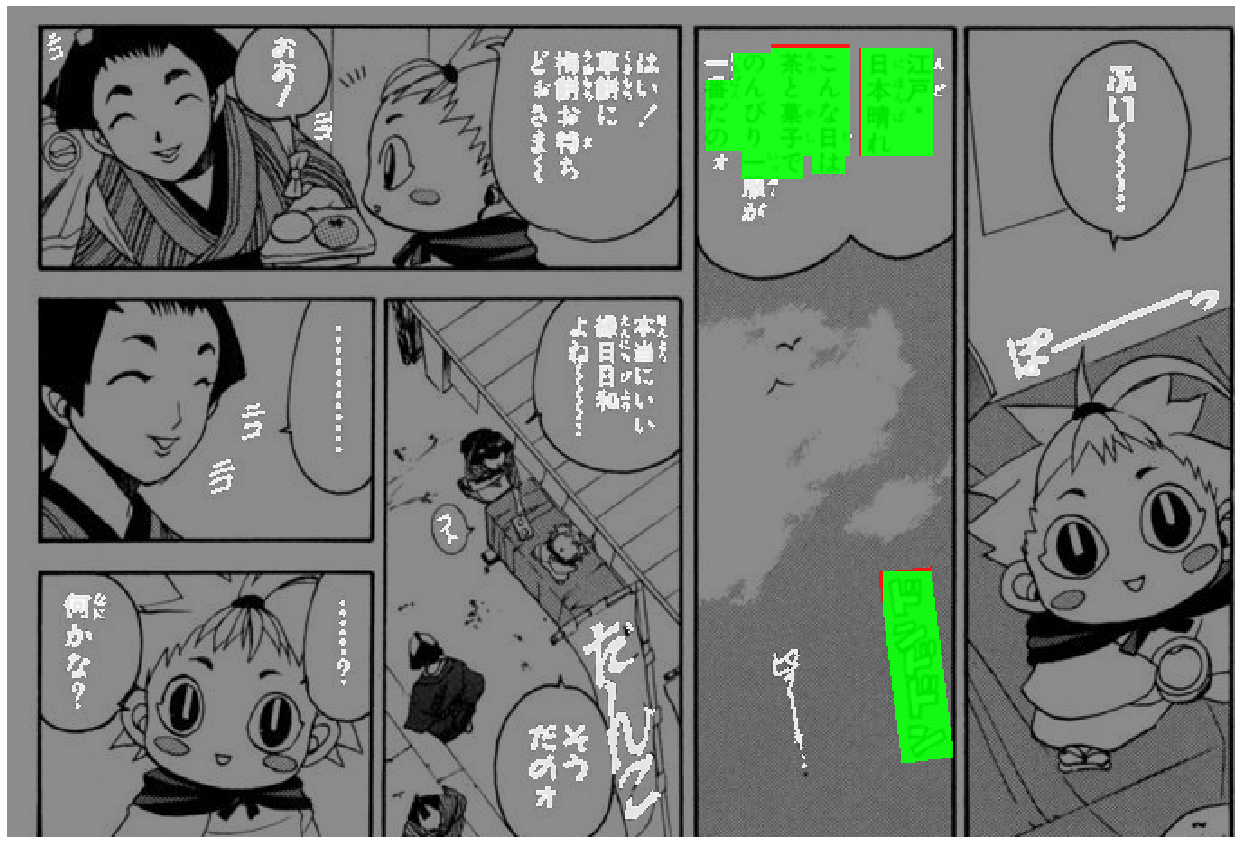}
    \caption{CRAFT}
\end{subfigure}
\begin{subfigure}[t]{\textwidth}
    \includegraphics[width=\textwidth]{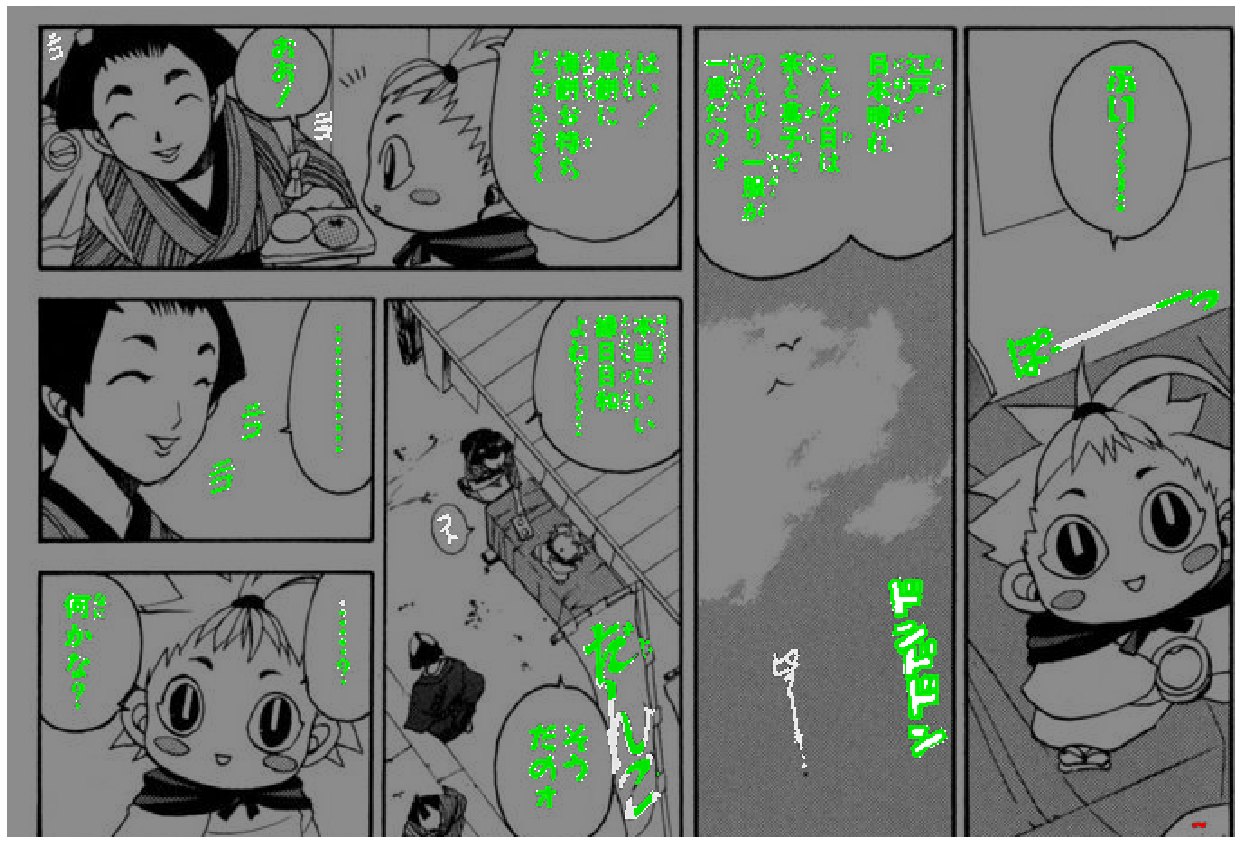}
    \caption{Ours}
\end{subfigure}
\caption{Segmentation masks obtained by CRAFT and our method. In white, missing text. In green, text correctly segmented (relaxed mode). Manga image extracted from the \texttt{Manga109} dataset \cite{manga109} \cite{akkera}. (a) Segmentation mask obtained by CRAFT. (b) Segmentation mask obtained by our method
}
\label{fig:craft}
\end{figure}

As CRAFT method outputs non-rigid bounding boxes, to make the comparison fair, we extend the dilation of the ground truth masks in the dataset to include the bounding box of each text connected component. We used the CRAFT pre-trained model provided by the authors \cite{craftgithub} and calculated the metrics for our dataset.
As can be seen in Tables \ref{table:similar_works} and \ref{table:similar_works_craft}, our model outperformed the CRAFT method in all metrics. An example can be observed in Fig. \ref{fig:craft}

\begin{table}[h!]
\centering
 \caption{CRAFT metrics for our dataset}
\begin{tabular}{ |c|c|c|c|c|  }
 \hline
 \multirow{2}{*}{Method} & \multicolumn{2}{c|}{Normal} & \multicolumn{2}{c|}{Relaxed} \\
 \cline{2-5}
 & PF1 & GF1 & PF1 & GF1 \\
 \hline
CRAFT & $45.18 \pm 2.8$ & $43.40 \pm 1.3$ & $73.72 \pm 5.2$ & $78.55 \pm 1.0$ \\
 \hline
\end{tabular}
\label{table:similar_works_craft}
\end{table}

\clearpage

\FloatBarrier
\subsection{Robustness}

\begin{figure}[h]
\centering
    \includegraphics[height=\textheight]{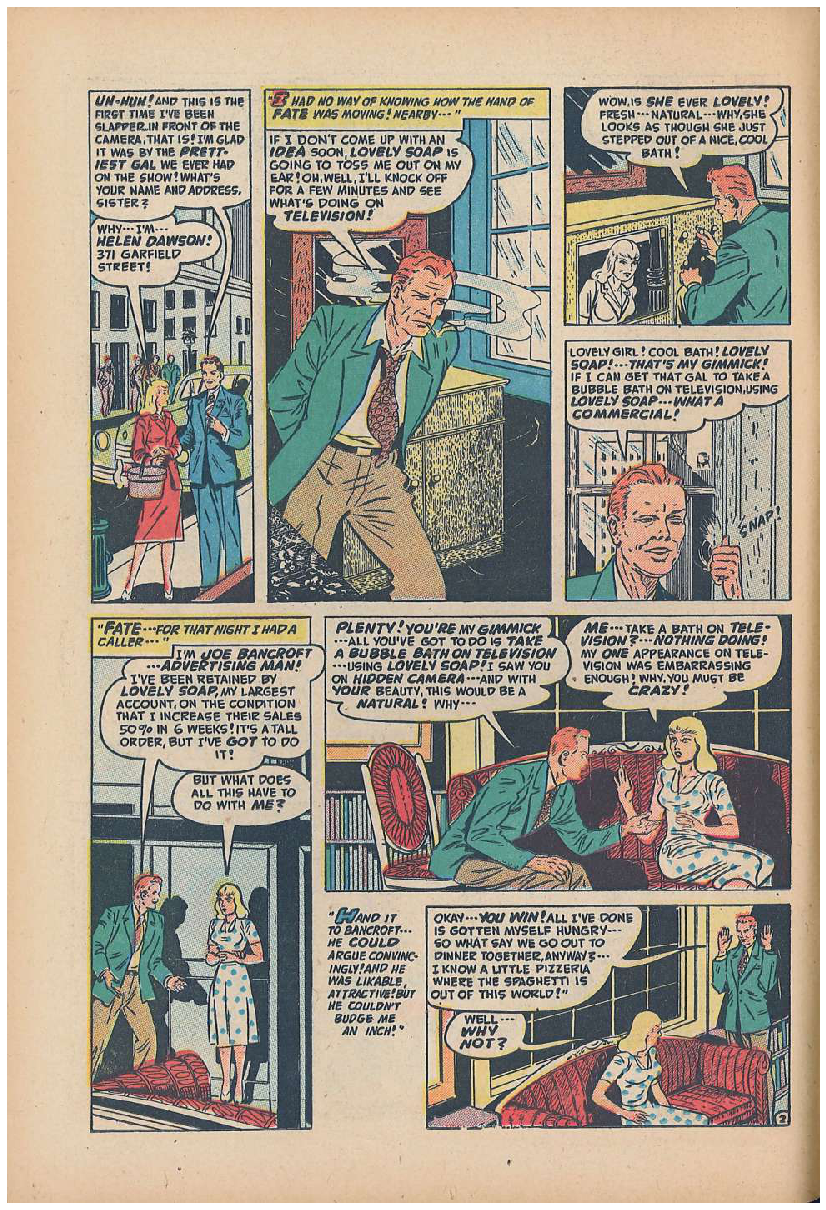}
    \caption{Example of a colour comic, the style is quite different to manga but it still has similar speech bubbles}
\label{fig:comic-segmentation-1}
\end{figure}

\begin{figure}[h]
\centering
    \includegraphics[height=\textheight]{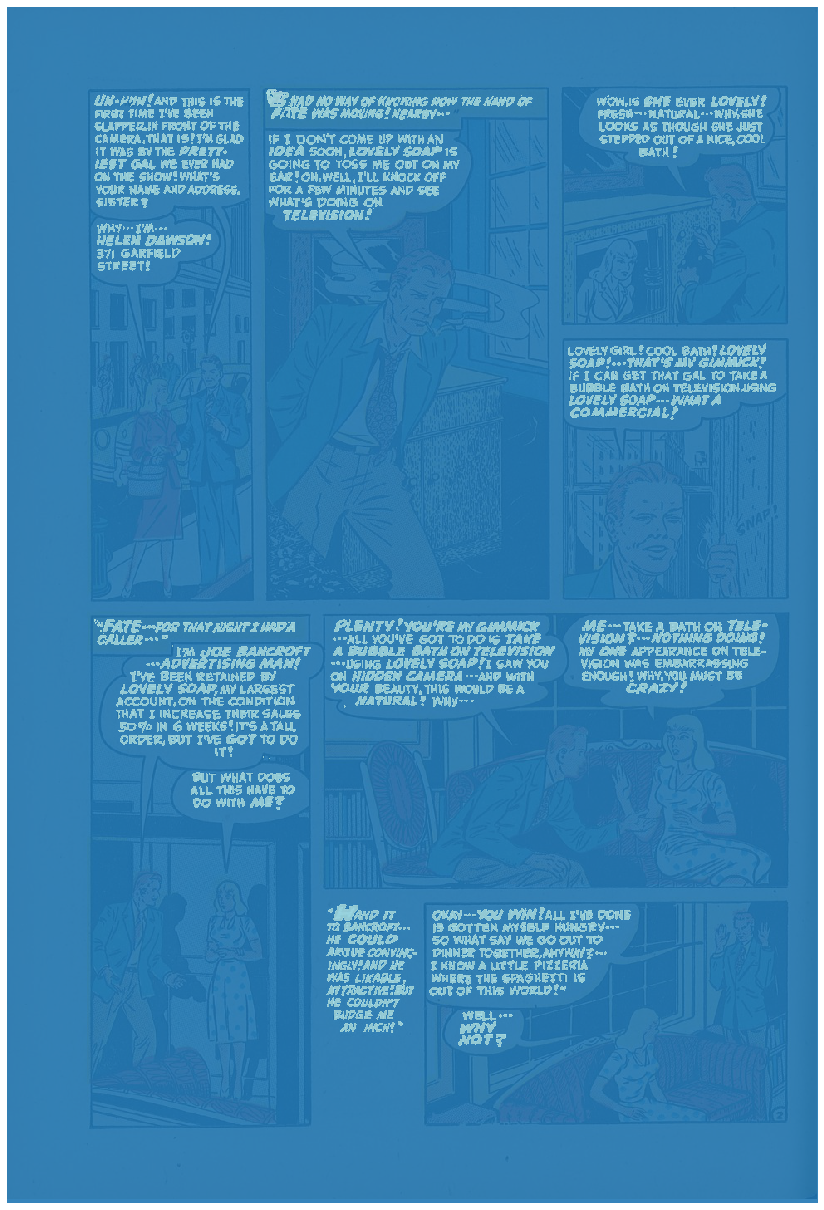}
    \caption{Example of prediction over a colour comic}
\label{fig:comic-segmentation-2}
\end{figure}

\begin{figure}[h]
\centering
    \includegraphics[height=0.45\textheight]{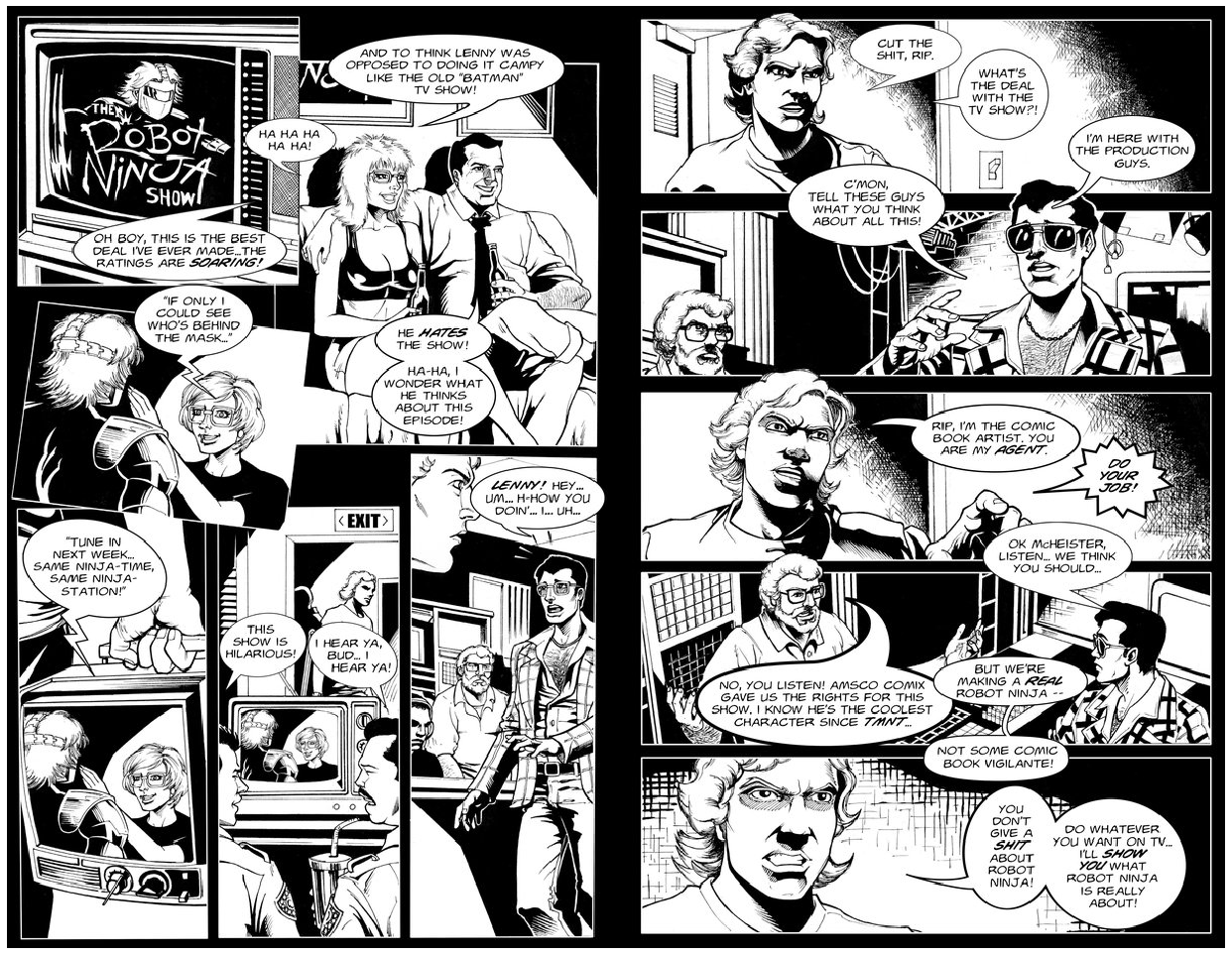}
    \caption{Example of a black and white comic}
\label{fig:comic-segmentation-3}
\end{figure}

\begin{figure}[h]
\centering
    \includegraphics[height=0.45\textheight]{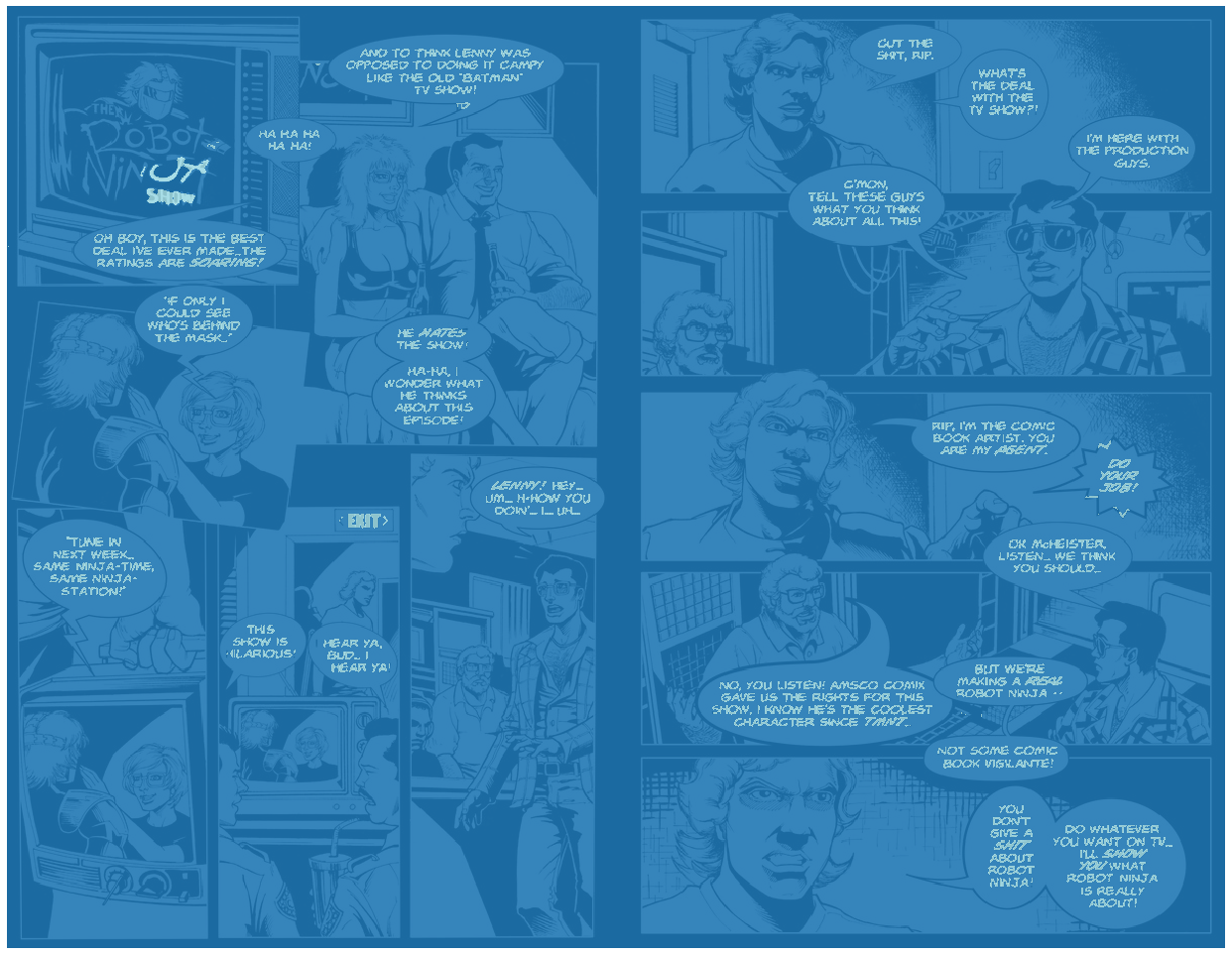}
    \caption{Example of prediction over a black and white comic}
\label{fig:comic-segmentation-4}
\end{figure}

We also did some preliminary experiments for predicting languages and genres not used during training. An example of a colour comic can be seen in Figs. \ref{fig:comic-segmentation-1}, \ref{fig:comic-segmentation-2}. Except for the snap word outside the speech bubbles, all text was correctly segmented with almost no false positives. An example of a black and white comic can be seen in Figs. \ref{fig:comic-segmentation-3}, \ref{fig:comic-segmentation-4}. The results were also very good.

\begin{figure}[h]
\centering
\begin{subfigure}[t]{0.49\textwidth}
    \includegraphics[height=0.45\textheight]{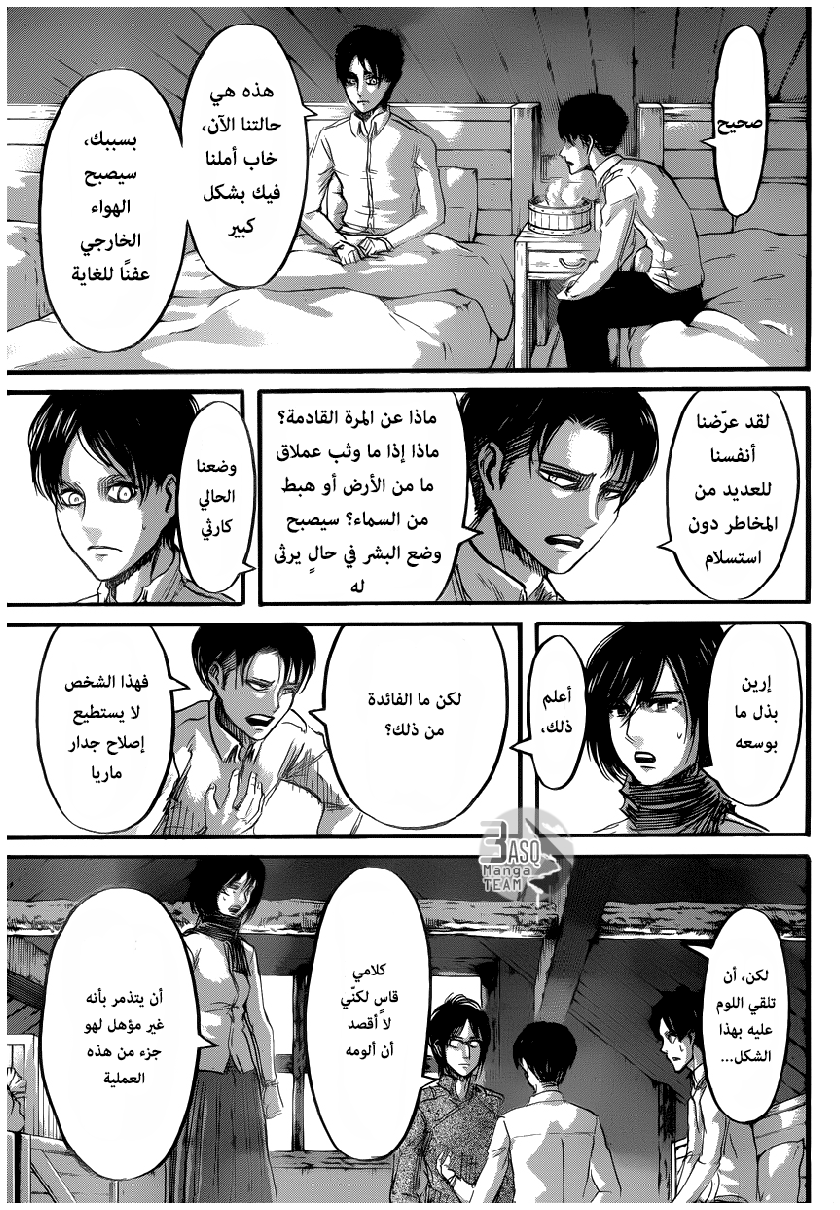}
\end{subfigure}
\begin{subfigure}[t]{0.49\textwidth}
    \includegraphics[height=0.45\textheight]{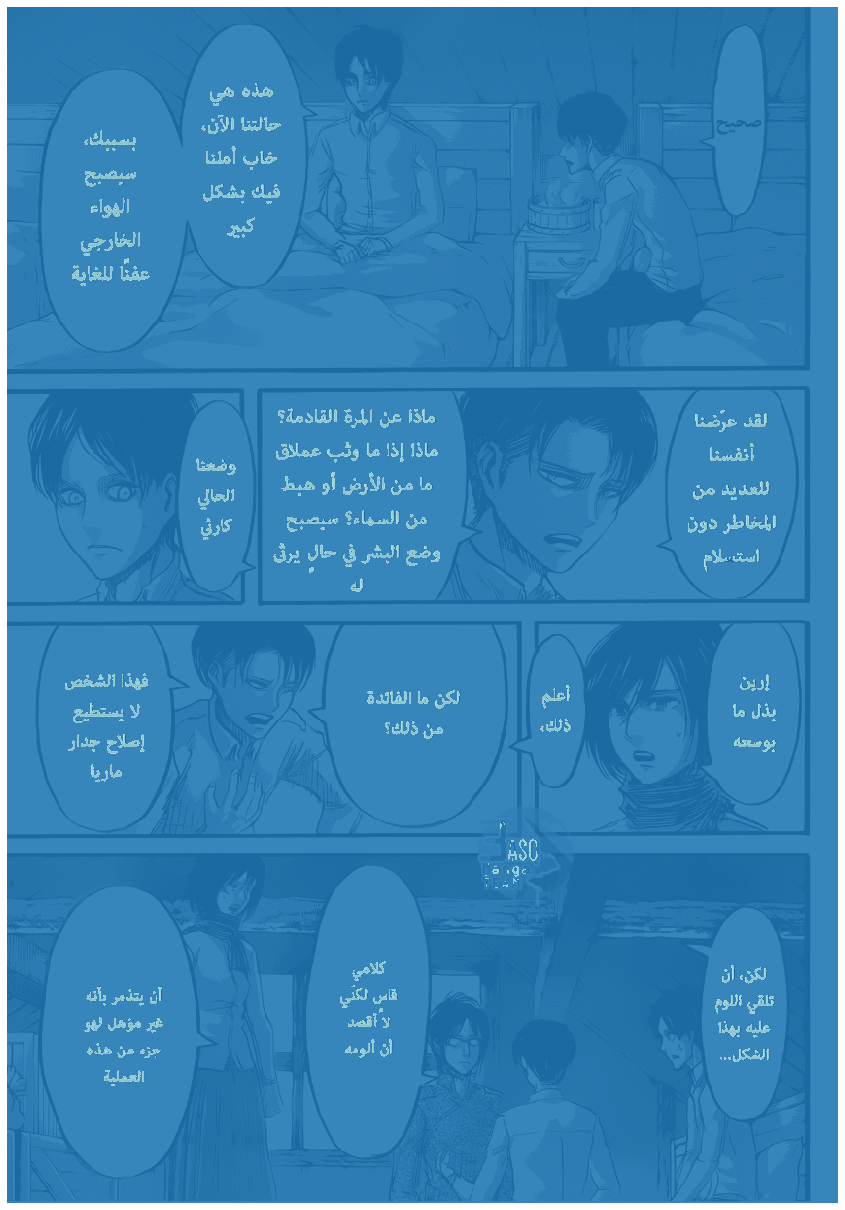}
\end{subfigure}
\caption{Example of a manga translated to arabic, (a) original image and (b) predicted mask}
    \label{fig:comic-segmentation-5}
\end{figure}

\begin{figure}[h]
\centering
    \includegraphics[height=0.35\textheight]{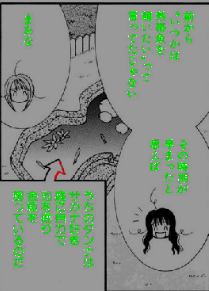}
    \caption{Segmentation mask obtained by our model for a manga image that has speech bubbles with both text and non-text. In red, false positives. In white, missing text. In green, text correctly segmented. 
    }
\label{fig:generalization}
\end{figure}

We also tried predicting the text of some translated mangas in Arabic, a language whose letters were never observed during training. While not as good as japanese manga, our model detected many of the Arabic letters inside the speech bubbles accurately, as seen in Fig. \ref{fig:comic-segmentation-5}.
One could come to the wrong conclussion and believe that the easy text has good predictions even if it was never observed in training because the network is merely predicting everything inside a speech bubble as text. However, we can see in Fig.  \ref{fig:generalization}  that our model does not predict the heads inside the speech bubbles as text.

\FloatBarrier
\subsection{Improvement over synthetic data}
To check how we improved from our initial idea of generating synthetic data against using real labeled data, we calculate our new metrics with the segmentation model trained from the danbooru images with generated text using 0.95 as threshold, the expansion algorithm and finally the remove noise algorithm. Results can be seen in Table \ref{table:synthetic}. We can see that going through the trouble of labelling manga images was worth it, there is a big difference in all metrics.

\begin{table}[h!]
\centering
 \caption{Synthetic metrics on our dataset}
\begin{tabular}{ |c|c|c|c|c|  }
 \hline
 \multirow{2}{*}{Method} & \multicolumn{2}{c|}{Normal} & \multicolumn{2}{c|}{Relaxed} \\
 \cline{2-5}
 & PF1 & GF1 & PF1 & GF1 \\
 \hline
Synthetic   & $45.11 \pm 4.1$ & $65.70 \pm 3.7$ & $34.09 \pm 4.3$ & $68.96 \pm 4.0$ \\
Real Data & \bm{$79.36 \pm 1.1$} & \bm{$84.92 \pm 1.0$} & \bm{$80.43 \pm 1.8$} & \bm{$89.26 \pm 1.0$} \\
 \hline
\end{tabular}
\label{table:synthetic}
\end{table}

\chapter{Conclusions}\label{conclusions}

The detection and recognition of unconstrained text is an open problem in research. Japanese Optical Character Recognition is also still a developing field. Standard methods developed for the Latin alphabet do not perform well with Japanese, due to Japanese having many more characters: about 2,800 common characters out of a total set of more than 50,000. Each Japanese character is, on average, more complicated than an English letter \cite{Basich2016OpticalCR}. 
Japan is a country with an immense cultural heritage. Unfortunately, the complexity of the Japanese language constitutes a linguistic barrier for accessing its culture. Automatic translation methods would contribute to overcome it.

In this work, we presented a study into unconstrained text segmentation at a pixel level in Japanese manga. We show our ideas and findings over different ways of handling the problem. We created a dataset manually annotating manga images and implemented special metrics to evaluate this task. We show that these tools, together with the  \texttt{fastai} library, allowed us to find a simple and efficient deep learning model that outperforms in most metrics previous works on the same task. Some preliminary experiments show that our model has good generalization, and is also robust for text detection inside speech bubbles for languages and comic genres not observed during training. The text segmentation masks obtained by our method could be useful for Japanese OCR and inpainting.
Lastly, the dataset and metrics provided by this work would enable other researchers and practitioners to find better models for this problem.

With the release of the ground truth and our predictions, it would also allow others to calculate other metrics we haven't mentioned or explored, such as PSNR (Peak Signal to Noise Ratio) or DRD (Distance Reciprocal Distortion). 

\bibliographystyle{plain}
\bibliography{egbib}

\end{document}